\documentclass[10pt,twocolumn,letterpaper]{article}

\usepackage[pagenumbers]{cvpr} 

\usepackage[dvipsnames]{xcolor}

\definecolor{cvprblue}{rgb}{0.21,0.49,0.74}
\usepackage[pagebackref,breaklinks,colorlinks,citecolor=cvprblue]{hyperref}

\def\paperID{16869} 
\def\confName{CVPR}
\def\confYear{2024}

\usepackage[utf8]{inputenc} 
\usepackage[T1]{fontenc}    
\usepackage{hyperref}       
\usepackage{url}            
\usepackage{booktabs}       
\usepackage{amsfonts}       
\usepackage{nicefrac}       
\usepackage{microtype}      
\usepackage{xcolor}         

\usepackage{amsmath,amssymb,amsfonts}
\usepackage{graphicx}
\usepackage{textcomp}
\usepackage{xcolor}

\usepackage{balance}

\usepackage{verbatim}
\usepackage{wrapfig}
\usepackage[utf8]{inputenc} 
\usepackage[T1]{fontenc}    
\usepackage{hyperref}       
\usepackage{url}            
\usepackage{booktabs}       
\usepackage{amsfonts}       
\usepackage{nicefrac}       
\usepackage{microtype}      
\usepackage{mathrsfs}
\usepackage{arydshln}

\usepackage{verbatim}
\usepackage{wrapfig}
\usepackage{soul}
\usepackage{wrapfig}

\usepackage{xcolor}
\usepackage{color,colortbl}
\definecolor{Gray}{gray}{0.8}
\usepackage{graphicx}

\usepackage{hyperref}
\usepackage{subcaption}
\usepackage{pifont}
\usepackage{tikz}
\usepackage{amssymb}
\usepackage{enumitem}
\setlist{nolistsep}

\usepackage{amsmath}
\usepackage{amsthm}
\usepackage{amssymb} 

\usepackage{mdframed} 
\usepackage{thmtools}

\usepackage{mathabx} 

\usepackage{graphicx}
\newcommand{\greenup}{\textcolor{green}{$\uparrow$}}
\newcommand{\reddown}{\textcolor{red}{$\downarrow$}}

\theoremstyle{definition}

\theoremstyle{remark}

\usepackage{mdframed} 
\usepackage{thmtools}
\usepackage{multirow}
\usepackage{mdframed} 
\newcommand{\myNum}[1]{(\emph{#1})}

\newcommand{\smartparagraph}[1]{\vspace{2pt} \noindent {\bf #1}}

\title{Multiview Aerial Visual Recognition (MAVREC): \\ Can Multi-view Improve Aerial Visual Perception?}

\author{Aritra Dutta$^{1}$, Srijan Das$^{2}$, Jacob Nielsen$^{3}$, Rajatsubhra Chakraborty$^{2}$, and Mubarak Shah$^{4}$\\
$^{1}$ AI Initiative, UCF, 
$^{2}$ UNC Charlotte, 
$^{3}$ IMADA, SDU,
$^{4}$ CRCV, UCF\\
{\tt\small aritra.dutta@ucf.edu} \hspace{0.1in}{\tt\small sdas24@charlotte.edu}\\
}

\begin{document}
\twocolumn[{%
\renewcommand\twocolumn[1][]{#1}%
\maketitle
\centering
\scalebox{1.0}{
\includegraphics[width=1\textwidth]{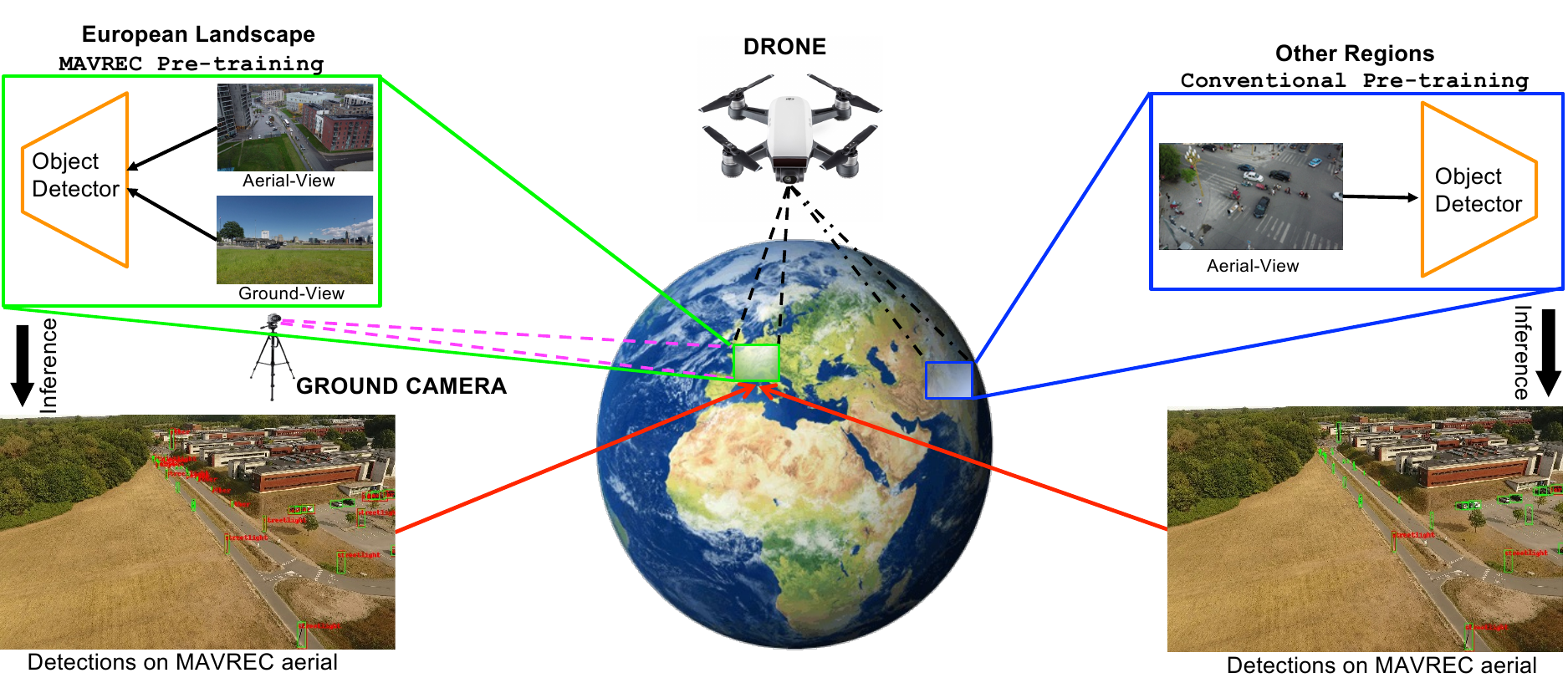}}
\captionof{figure}{Illustration of the geography-aware model using our proposed MAVREC dataset (\textcolor{green}{green} box) collected in the rural and urban European landscape vs. the conventional aerial object detector (\textcolor{blue}{blue} box) pretrained only on aerial images from VisDrone \cite{visdrone} captured in Asia. The conventional approach fails to precisely detect aerial objects from the MAVREC dataset. In contrast, our object detector pretrained on the ground and aerial images from the MAVREC dataset contextualize the object proposals of that specific geography and enhance the aerial visual perception, thus outperforming other object detectors pre-trained on popular ground-view dataset (MS-COCO~\cite{COCO}) or other aerial datasets collected from different geographies; also, see Figure \ref{fig:teaser}.}\label{fig:TEASER-world}}\vspace{2ex}]


\begin{abstract}
Despite the commercial abundance of UAVs, aerial data acquisition remains challenging, and the existing Asia and North America-centric open-source UAV datasets are small-scale or low-resolution and lack diversity in scene contextuality. Additionally, the color content of the scenes, solar-zenith angle, and population density of different geographies influence the data diversity. These two factors conjointly render suboptimal aerial-visual perception of the deep neural network (DNN) models trained primarily on the ground-view data, including the open-world foundational models. 

To pave the way for a transformative era of aerial detection, we present \textbf{M}ultiview \textbf{A}erial \textbf{V}isual \textbf{REC}ognition or \textbf{MAVREC}, a video dataset where we record synchronized scenes from different perspectives --- ground camera and drone-mounted camera. MAVREC consists of around 2.5 hours of industry-standard 2.7K resolution video sequences, more than 0.5 million frames, and 1.1 million annotated bounding boxes.~This makes MAVREC the largest ground and aerial-view dataset, and the {fourth} largest among all drone-based datasets across all modalities and tasks.~Through our extensive benchmarking on MAVREC, we recognize that augmenting object detectors with ground-view images from the corresponding geographical location is a superior pre-training strategy for aerial detection. Building on this strategy, we benchmark MAVREC with a curriculum-based semi-supervised object detection approach that leverages labeled (ground and aerial) and unlabeled (only aerial) images to enhance the aerial detection. We publicly release the MAVREC dataset: \url{https://mavrec.github.io}.





\end{abstract}  
\section{Introduction}\label{sec:intro}

Object detection and tracking employing UAV~(or drone)-based aerial videos are essential in many downstream applications, such as autonomous driving \cite{Marathe_2022_CVPR, bojarski2016end}, robotics \cite{wu2022_survey}, environmental monitoring \cite{PanETAL2021}, infrastructure inspection \cite{ASK2020}, developing livable and safe communities \cite{Islam2023EffectOS, cityscape, UCF-SST-CitySim}, a few to name. Despite having many crucial applications, most visual perception models focus on ground-view images. This bias results in suboptimal performance when these models are applied to an aerial perspective --- a discrepancy due to a domain shift precipitated by the viewpoint transfer. We hypothesize that the basis of this disparity is primarily twofold:

{First, the lack of diversity in the current aerial datasets.} Modern DNN-based visual models are data-hungry. However, aerial data collection is intricate due to UAV flight regulations and safety protocol, atmospheric turbulence, and many more~\cite{kakaletsis2021computer}. The existing open-source UAV datasets \cite{visdrone, UAVid, MOR-UAV, Au-Air, UAV123, UAVDT, DroneVehicle, BIRDSAI} are either small-scale, or low-resolution, and collected primarily in the urban pasture across Asian and North American geographies. These factors contribute to inadequacy in diverse dataset properties and hinder training large DNNs for aerial visual perception. 

Second, {substandard generalizability of the existing aerial visual models across different geographic locations.}~An object detection model trained on datasets captured from South Asia underperforms when deployed on videos captured with European landscapes, characterized by semi-rural pastures and lots of greenery; see Figure \ref{fig:TEASER-world}.~Research shows that 
different geographic factors, including latitude influences the population density \cite{kummu2011world, latitude, kummu2016over}, hence, the {\em color-content of the scenes}\footnote{European vehicles are comprising of mainly three colors \cite{car-color-2}; also, see \ref{sec:Appendix-color} for an analysis.},~their complexities, and density and interactions of the foreground objects. These seemingly low-key factors directly affect the data captured from a drone-mounted camera, and the previous studies never considered the inter-domain inference quality of the DNN models trained on these data.

To pave the way for a transformative era of aerial visual perception models, we introduce \textbf{M}ultiview \textbf{A}erial \textbf{V}isual \textbf{REC}ognition dataset, \textbf{MAVREC}, which uniquely captures time-synchronized aerial and ground view data.~MAVREC is collected with consumer-grade handheld cameras~(smartphones and GoPro)~and drone-mounted cameras, consists of around 2.5 hours of industry-standard 2.7K resolution video sequences, more than 0.5 million frames, covering rural and urban pastures during spring and summer in high-latitude European geographies.~It makes {\em MAVREC the largest ground and aerial-view dataset}, and {\em the {fourth} largest among all drone-based datasets {(edited and unedited)} across all modalities and tasks that ever existed}; see Table \ref{tab:summary}. 

In this paper, we rigorously assess our hypothesis and explore interesting properties of object detection in aerial images while evaluating MAVREC in a supervised setting. We find that contextual information of the landscape vastly influences aerial object detection, which is not the case for general object detection in ground view. Therefore, when an aerial object detection model is trained exclusively within a specific geographical context, it often exhibits limited generalizability across diverse geographical locations. This limitation necessitates the development of aerial detection models capable of understanding and integrating geography-specific features.
In our experimental analysis with MAVREC, we find that transfer learning from ground to aerial view induces geography-aware representations in aerial object detection models. This curriculum-based training approach notably surpasses the performance of object detectors pre-trained on alternative aerial or ground datasets, including advanced foundational models such as Grounding DINO~\cite{liu2023grounding}, trained on extensive data corpora; see details in \S\ref{sec:evaluation}.
Furthermore, we reckon significant resource investment in annotating large-scale aerial object detection datasets. Annotations are cost-intensive and demand substantial human intervention and time, rendering them impractical for numerous real-world applications. To this end, we benchmark MAVREC with a curriculum-based semi-supervised object detection approach that leverages labeled and unlabeled images to enhance the detection performance from an aerial perspective.

We summarize our key technical contributions as follows: 
\begin{itemize}
    \item We introduce MAVREC, which to date represents the most extensive dataset integrating time-synchronized ground and aerial images captured in the European landscape; \S\ref{sec:dataset}.
    \item Through benchmarking MAVREC in supervised and semi-supervised settings, we expose the proclivity of existing pre-trained object detectors to exhibit bias toward data captured from ground perspectives; \S\ref{sec:evaluation}. 
    \item We propose a curriculum-based semi-supervised object detection method.~Its superior performance shows the importance of training these types of models with ground-view images to learn geography-aware representation; \S\ref{sec:semi-supervised-benchmark}.
\end{itemize}

\begin{table*}[t!] 
    \caption{{State-of-the-art UAV-based datasets since 2016 in chronological order.~For viewpoints, G denotes {\it ground-view}, A denotes {\it aerial-view}, and AG denotes both. 
    Thermal IR datasets are not included.}}\label{tab:summary}
    \small
    \centering
    \scalebox{0.9}{
    \begin{tabular}{@{}lccccccccc@{}}
     \multicolumn{9}{c}{} \\
     \midrule
     \textbf{Dataset} & \textbf{Total}  & \textbf{Resolution} & \textbf{Total} & \textbf{{Instances per}}  & \textbf{Categories} & \textbf{Viewpoints} & \textbf{Region}  & 
     \textbf{Year}\\
             & \textbf{Frames} &            & \textbf{Annotations} & \textbf{{Annotated Frame}}      &            &              &        &      \\
    \midrule
     Campus \cite{campus} & 929,499 & $1400 \times 2019$  & 19,564 & 0.02 & 6  & Single (A) & North America & 2016 \\
     UAV123 \cite{UAV123}  & 110,000& $720 \times 720$  & 110,000 & 1.0 & 6  & Multi (A)& Middle East & 2016 \\
     CarFusion\cite{CarFusion}   &53,000 & $1,280 \times 720$  & --  & -- & 4  &Multi & North America &  2018 \\
     DAC-SDC \cite{DAC-SDC} & 150,000& $640 \times 360$  & NA & NA & 12 & Single & Asia & 2018\\
     UAVDT \cite{UAVDT}   & 80,000 & $1080 \times 540$   & 841,500 & 10.52 & 3 & Single & Asia & 2018 \\

     MDOT \cite{MDOT}   & 259,793 &  --  & --  & -- & 9 & Multi (A)& Asia & 2019 \\
     Visdrone DET \cite{visdrone}   & 10,209 & $3840 \times 2160$  & 471,266 & 53.09 & 10  & Single (A)& Asia  & 2019\\ 
     Visdrone MOT \cite{visdrone}   & 40,000 & $3840 \times 2160$  & 1,527,557 & 45.83 & 10  & Single (A)& Asia  & 2019\\ 
     DOTA\cite{xia2018dota} &  2806 & $4000 \times 4000$ & 188,282 & 67.09  & 15 & Single (A)& {Multiple} & 2019\\ 
     {DOTA V2.0~\cite{ding2021object}} &  11,268 & $4000 \times 4000$ & 1,793,658  & 159.18  & 18 & Single (A)& {Multiple} & 2021\\ 
     MOR-UAV \cite{MOR-UAV} & 10,948 & $1280 \times 720,$ & 89,783 & 8.20 & 2 & Single & Asia & 2020 \\
     & & $1920 \times 1080$ & & & & & \\
     AU-AIR \cite{Au-Air} & 32,823 & $1920 \times 1080$  & 132,034 & 4.02 & 8 & Multi & Europe & 2020\\
     UAVid \cite{UAVid}   & 300 & $3840 \times 2160$  & --  & -- & 8 &Single & Europe  & 2020 \\
     && $5472 \times 3078,$  &&&&&&\\
     MOHR \cite{mohr} & 10,631 & $7360 \times 4192,$ & 90,014 & 8.47  & 5 &  Multi (A) & Asia & 2021\\
     && $8688 \times 5792$ &&&&&&\\
     \midrule
    \rowcolor{Gray}
    \textbf{MAVREC}{~(This paper)} &537,030 & $2700 \times 1520$ & 1,102,604 & 50.01 & 10 & \textcolor{red}{\textbf{Multi (AG)}} & Europe & 2023 \\ 
    \midrule
    \end{tabular}}
    \end{table*}

\section{Related work}\label{sec:related_work}

\smartparagraph{UAV-based datasets.}~The last decade witnessed a surge in UAV-based video and image datasets.~We list some open-source UAV datasets, curated since 2016, and group their key features according to their downstream tasks. 

\textit{VisDrone} \cite{visdrone} is the most widely used drone dataset for aerial image object detection. It is recorded from 14 cities in China with various drone-mounted cameras, consists of 10 object categories, and segregated into four task-specific sub-datasets: (a) Image Object Detection (10,209 images), (b) Video Object Detection (96 videos, 40,001 images), (c) Single-Object Tracking (139,276 images), and (d) Multi-Object Tracking (40,000 images). 
\textit{Campus} \cite{campus}, is the largest aerial dataset for multi-target tracking, activity comprehension, and trajectory prediction, focuses solely on the university campus, in contrast to our MAVREC.
~\textit{UAVDT} \cite{UAVDT} dataset consists of 80,000 frames and 3 subsets, focusing on single and multi-object detection and tracking, under different weather condition, lighting, and altitude of the drone. \textit{MOR-UAV} \cite{MOR-UAV} is an aerial dataset consisting of 10,948 images, all annotated, designed for moving object detection under various challenges, such as illumination, camera movement, etc. \textit{UAV123} \cite{UAV123} is a low-altitude aerial dataset consisting of 112,578 fully-annotated images across 123 video sequences (simulated and recorded), designed for object tracking, with a subset intended for long-term aerial tracking. \textit{MDOT} \cite{MDOT} is a {\em multi-drone based single object tracking dataset} with 259,793 frames across 155 groups of video clips, and 10 different annotated attributes.
~\textit{Au-Air}~\cite{Au-Air} is a medium scale, multi-sensor, aerial data designed for real-time object detection, with the aim of bridging the gap between computer vision and robotics. \textit{DAC-SDC} \cite{DAC-SDC} is a single-object detection dataset with 150,000 images collected from \textit{DJI} \cite{dji} with 12 categories. {\em DOTA} \cite{xia2018dota} is an aerial dataset (2,806 images, 15 categories) for object detection in earth vision.~{{\em DOTA V2.0}~\cite{ding2021object}, an upgraded version of DOTA, is a single-view dataset collected from Google Earth, GF-2 satellite and aerial imagery (11,268 images, 18 categories) for object detection.~The $UG^2+$ Challenges provide \textit{A2I2-Haze} \cite{UG_two}, the first real haze dataset, consisting of 229 pairs of hazy and clean images (197 training pairs, 32 testing pairs) from 12 videos, focusing on detection in visually degraded environments with smoke and haze with mutually exclusive aerial and ground images.} 

In an orthogonal line of work,~\textit{GRACO}~\cite{GRACO} is a multimodal dataset for synchronized ground and aerial collaborative simultaneous localization and mapping (SLAM) algorithms (6 ground and 8 aerial sequences collected in China within a university campus) by a group of ground and aerial robots equipped with light detection and ranging (LiDAR), cameras, and global navigation satellite/inertial navigation systems~(GNSS/INS) that capture images at 20Hz with 2182.52 and 2675.54 seconds duration in the ground and aerial, respectively.~\textit{S3E}~\cite{feng2022s3e} is a multimodal dataset for collaborative SLAM, consists of 7 outdoor and 5 indoor synchronized ground sequences, each longer than 200 seconds, and collected in 5 locations within a university campus in China. \textit{DVCD18K}~\cite{ashtari2022drone} is a cinematographic dataset (with corresponding camera paths) consisting of 18,551 edited drone clips spanning 44.3 hours.

Our proposed \textit{MAVREC} is inherently different from the above datasets, because:~\myNum{a}~compared to other small-scale~(e.g.,~UAVid, DOTA,~MOR-UAV,~AU-AIR), and low-resolution datasets~(e.g.,~UAV123, UAVDT, DroneVehicle, BIRDSAI), MAVREC is the first-ever {\em large-scale}, unscripted, multi-viewpoint video dataset~(fourth largest among all UAV-based datasets ever after {DVCD18K, Highway-drone, and} Campus; {Campus} lacks object boundaries) recorded in {\em industry-standard} 2.7K resolution; \myNum{b} its multi-viewpoint presents the same scenes through the lens of one or more ground cameras, and a medium altitude~(flight height 25--45 meters, compared to low or high-altitude datasets, e.g., UAV123 with flight height 5--25 meters, MOHR with flight height 200 meters or above) drone-mounted camera~(this perspective is unique compared to the existing multiview drone datasets, e.g., MDOT, UAV123, MVDTD, MCL) to have balanced variations of small and medium objects from both perspectives; \myNum{c} its high variance in object distribution across different scenes is complementary to datasets like VisDrone where object detection is relatively straightforward due to their biased object distribution (dense), reflecting its demographic characteristics; \myNum{d} {{\em MAVREC is the first multi-view, drone-based dataset curated in the wild, with a central focus on object detection}.~GRACO and S3E are multi-view but confined to university campuses and used for SLAM algorithms; S3E does not incorporate drone-based data acquisition. Alongside, {A2I2-Haze} dataset a tiny subgroup of \cite{UG_two} challenges, consists of mutually exclusive, non-synchronized aerial and ground images, while DVCD18K is an {\em human-edited cinematographic dataset with drone camera paths}, and vastly different from MAVREC in many aspects.} \S \ref{sec:stat} explains more unique challenges of MAVREC. 
\begin{figure*}
    \centering
    \includegraphics[width=0.95\textwidth]{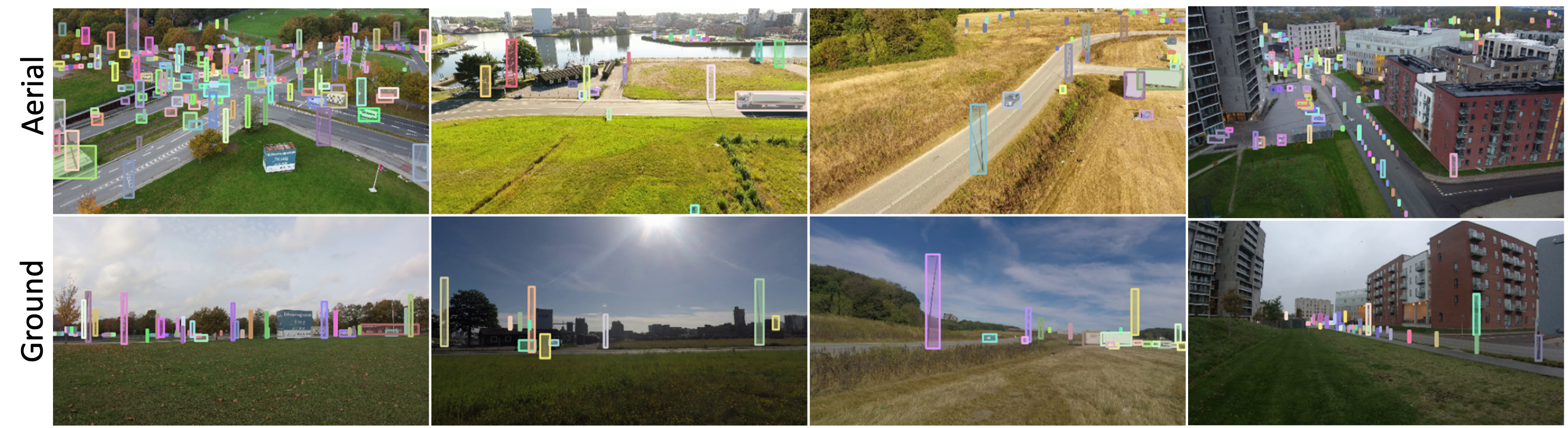}
    \caption{{Different sample scenes (with annotation) from our dataset; the first row is the aerial-view, second row presents the same scenes from a ground camera. See more sample frames in \S \ref{sec:Appendix-Dataset}, Figure \ref{fig:Different_scenes_from_the_dataset_extra}.}}\label{fig:Different_scenes_from_the_dataset}
\end{figure*}
There are UAV-based datasets with downstream tasks primarily orthogonal to MAVREC. For completeness, we list some UAV-based datasets for action detection, counting, geo-localization, 3D reconstruction, and benchmarking in \S \ref{app:relatedwork}; also, see \cite{wu2022_survey}. 

\myNum{iii}\smartparagraph{Object detection.} CNN-based object detectors are divided into two categories: two-stage and one-stage. Two-stage detectors such as RCNN~\cite{rcnn}, Fast RCNN~\cite{fast_rcnn}, Faster RCNN~\cite{fasterrcnn_humandetector}, employ a class-agnostic region proposal module followed by simultaneously regressing the object boundaries and their classes. In contrast, one-stage detectors like SSD~\cite{ssd}, YoloV4 \cite{yolov4}, YoloV6 \cite{yolov6}, YoloV7 \cite{yolov7}, YoloX \cite{yolox}, FCOS \cite{tian2020fcos}, directly predicts the image pixels as objects, leading to models that offer fast inference. Recently, by using neural architecture search, Yolo-NAS \cite{yolonas} outperforms previous Yolo models in real-time object detection. 
With the success of transformers, DETR \cite{DETR} was the first transformer-based, end-to-end object detector. Following this, Deformable-DETR~(D-DETR) \cite{D-DETR} introduces a sparse attention module, computationally $6\times$ faster than DETR, and robust in detecting small objects. The majority of object detectors designed for aerial imagery draw upon the foundational principles established by these aforementioned popular object detectors \cite{Xu_2022, xu2022rfla, WANG2023104697}. Along this line, TPH-YoloV5 \cite{zhu2021tph} combines YoloV5 with a transformer prediction head to solve the varying object scales and motion blur for drone-captured scenarios. As a result, our analysis utilizes the MAVREC dataset to benchmark these well-established methods, prioritizing factors such as fast inference, high precision, and the effective detection of small-scale objects.


\section{MAVREC dataset}\label{sec:dataset}

In this section, we start with the data acquisition process; and then explain annotation, statistical attributes, and unique challenges of MAVREC. 
\subsection{General setup}\label{sec:setup}
\smartparagraph{Recording set-up.} {We conduct the recording in public spaces in compliance with the European Union's drone safety and Scandinavian video surveillance regulations; see \S \ref{sec:appendix_impact} for detailed discussion on reproducibility, licensing, privacy, safety, maintenance plan and broader impact of the dataset.} {We record our 
dual-view aerial-ground dataset with a drone-mounted camera (DJI Phantom 4, DJI mini 2) and a consumer-grade static ground camera (GoPro Hero 4, GoPro Hero 6, iPhone 11 and 13-Pro) placed on a tripod; see details 
in Table \ref{tab:hardware_specification}. The drone is kept semi-static, hovering approximately 25--45 meters above the ground; see the relative positions and viewing angles of the drone and the ground camera in Figure \ref{fig:recording_setup_stat}.~Based on that,
we identify three recording scenarios (P1, P2, and P3).~
In P3, we better capture the objects as the drone gets a wider viewing angle.~However, we keep all views {\it not to amplify biases} from any particular view. For some recordings in the city center, railroad, or crowded intersections, we were unable to operate a drone due to the UAV-flight regulations; hence, we used a user-grade handheld camera set-up in the balcony of a high-riser to capture aerial views. 
\begin{figure*}[t]
     \centering
     \scalebox{1}{
     \includegraphics[width=\textwidth]{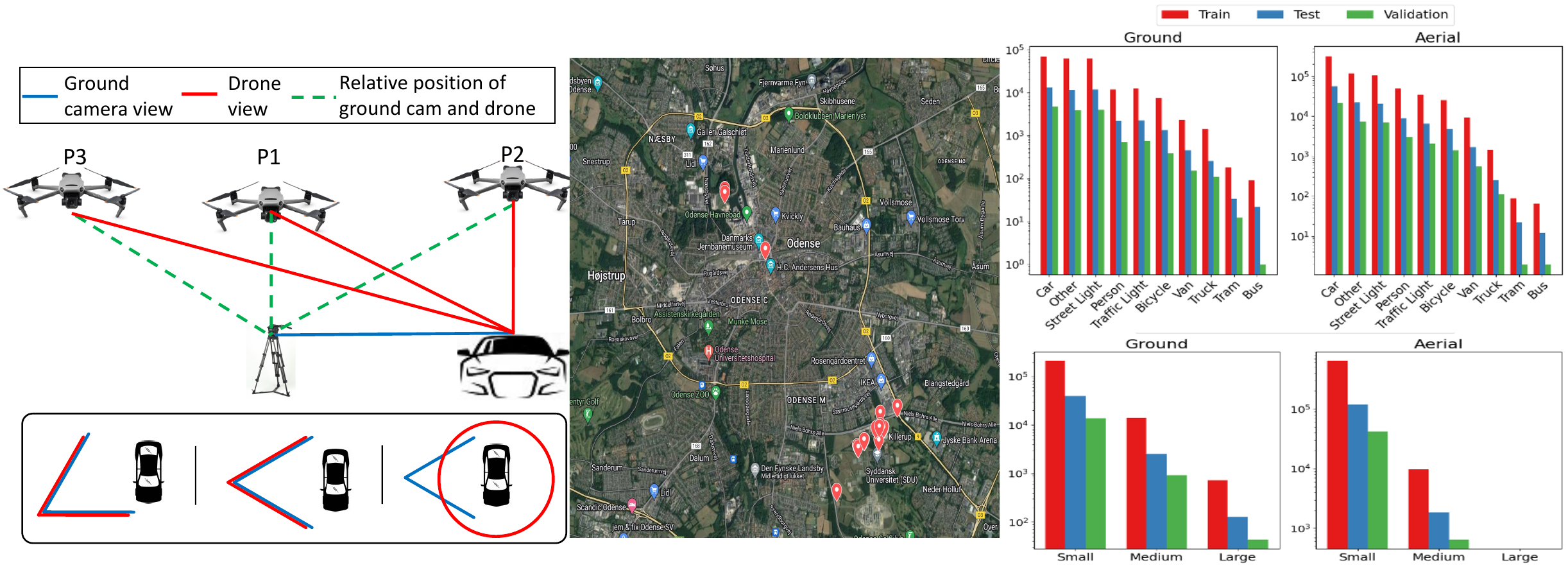}}
     \caption{{\textbf {Left}: Recording instances are divided into three different scenarios (P1, P2, and P3) based on the relative positions and the field-of-view (FOV) of the ground camera and the drone. The drone operates directly on top of the object (P2), and two oblique views---directly on top of the ground camera (P1), and behind the ground camera (P3). \textbf {Middle}: Recording locations as \textcolor{red}{red} dropped pins on the google map's sattelite view. \textbf{Right (top)}: Total numbers of objects in each category in the ground and aerial view. \textbf{Right (bottom)}: Number of large, medium, and small objects in the test, train and validation sets of two views; aerial view has no {\em large} object annotation.}}
     \label{fig:recording_setup_stat}
 \end{figure*}

\smartparagraph{Recording locations and scenes.} To avoid locational bias, we collected our data in 11 different geographical locations (European outdoors, rural and urban) with mixed pastures, in spring and summer (with the sun hitting the cameras from different angels), and when there is an encyclopedic spectrum of green and yellow intertwined in the background; see Figures \ref{fig:recording_setup_stat} and \ref{fig:Different_scenes_from_the_dataset} (also, see \ref{sec:Appendix-color} for an analysis). We choose the parking lots, and busy traffic intersections in the city, during the peak traffic hours to create more nuanced and complex interactions, in which multiple foreground objects are interacting and creating enormous visual challenges. Alongside, we choose harbor, single-lane roads in the countryside, asphalt roads, and bi-cycle lanes, in moderate traffic conditions, to collect simple scenarios which might have sparse to dense foreground objects (see sample frames in Figure \ref{fig:Different_scenes_from_the_dataset}). 

\smartparagraph{Alignment of dual-views.} Human operators simultaneously record the scenes from dual views; although a minute time-lapse is unavoidable. Consequently, after recording, clips are loaded into the {\tt QuickTime} player, and a human operator manually synchronizes the frames to alleviate the time lapses. We note that for 12.5\% of the clips (22.3\% of image frames), we captured an extra ground view.~Thus, these video clips offer three perspectives in total. 
{The inclusion of a second ground camera aims to enhance ground-to-ground representation. Although these videos have not been utilized in this paper, they are preserved for 
future work. }


\smartparagraph{Annotation and categories.}~MAVREC, as highlighted previously, stands as one of~the largest drone datasets, encompassing millions of objects within its distribution. However, annotating each object 
is a resource-intensive task. 
Inspired by the recent success of the semi-supervised learning paradigm in the computer vision community~\cite{OMNI-DETR, chen2022dense,liu2022unbiased, wang2023consistentteacher, mi2023active, Li_2022_CVPR}, we split the videos from different views into two categories; an annotated set and an unannotated set. After pre-processing, we select the first 30 seconds of the synchronized videos and annotate the frames through a semi-automatic, open-source annotation platform by Intel, called {\tt CVAT} \cite{CVAT}, and leave the rest of the video unannotated; see {\tt CVAT} interface in Figure \ref{fig:CVAT}. We provide an annotation interface with 10 categories in {\tt CVAT}: \texttt{tram}, \texttt{truck}, \texttt{bus}, \texttt{van}, \texttt{car}, \texttt{bicycle}, \texttt{person}, \texttt{street light}, \texttt{traffic light}, and \texttt{other}. In  \texttt{other} category, we annotated objects that share visual similarities with objects from the remaining categories; e.g., blocks of concrete from aerial view might look like cars, or white divider and marker posts from aerial view might look like a person with a white T-shirt, and so on. We created this category for the models to learn to disambiguate the {\em look-alike} objects from different categories. The in-built tracker in {\tt CVAT} tracks an object through multiple frames. We annotated by skipping forward 10 frames; thus speeding up the annotation process. {Nevertheless, to ensure high annotation quality, a human annotator reviews each frame 
and 5 human non-annotator have reviewed the dataset annotations. The annotation reviewers check that the bounding box encapsulates the object or its parts, has minimal overlap with other objects, and that all instances of the class in the frame are labelled. This process is performed on 600 annotated images randomly samples from the annotated data. During the review process, we find that the error is around 6\% which is comparable to the benchmark datasets in this domain.} Similar to other benchmarks \cite{visdrone,COCO}, annotated frames are assembled into {\tt COCO-json} format to give a unique identifier for each object class.


\subsection{Structuring, statistics, and challenges}
\label{sec:stat}

This section discusses the size, statistical properties, and challenges of the training distribution of MAVREC. We focus on two key points: \myNum{i} distribution of different categories, and \myNum{ii} distribution of the annotated object size.

\smartparagraph{Structuring the dataset.} 
We divided the annotated data from both views into three subsets---train, validation, and test sets. To ensure the distributions of the different objects are approximately the same throughout these three sets, we split each video sequence into three fragments, and then randomly select samples for each set. 

\smartparagraph{Distribution of different categories.} We show the distribution of categories from both views in Figure \ref{fig:recording_setup_stat}; also, see Figure \ref{fig:class_distribution-appendix}. MAVREC contains over {1.1 million bounding box annotations in both views combined, rendering $\sim50.01$ annotations per frame}; see Figure \ref{fig:recording_setup_stat} and details in Table \ref{tab:annotation_summary}. The distribution is {\em long-tailed} where cars are more frequent than trams and buses. The slight inconsistency in the object distribution from both views is natural as some recordings were conducted with the P3 setup, and in this setup, the drone has a wider viewing angle than the ground camera.

\smartparagraph{Object size distribution.} 
To better illustrate the challenges in MAVREC, we divide the object sizes present in the videos into {\em three} categories:~small ($<32\times32$ pixels), medium (lies inclusively between $32\times32$ and $96\times96$ pixels), and large ($>96\times96$ pixels). Figure~\ref{fig:recording_setup_stat}~(also, see Figure~\ref{fig:object_size_distribution}) presents the number of annotated object sizes in both views.~Large objects, such as trams, buses, and trucks, are present in fewer frames compared to the other objects. 
Also, the drone is maneuvered at a higher altitude, and the aerial view has a higher percentage of small objects compared to the ground view, creating a natural bias in object sizes. We also observe that the distribution for the split into the train, validation, and test set has almost an equal distribution of the different object sizes for both views; see Figure \ref{fig:object_size_distribution}-(c) for distribution for the object sizes.

\smartparagraph{Unique properties of MAVREC.}
MAVREC contains typical outdoor activities characterized by real-world properties like long-tail distribution, objects with similar appearance, viewpoint changes, varying illumination, etc. Additionally, MAVREC exhibits some unique properties, not found in other datasets:~\myNum{i} Ground view contains occluded objects. Nevertheless, these objects can be recovered due to the wide aerial FOV. This \textit{dual-view feature of the MAVREC} has the potential to offer a wide range of solutions for scenes with occlusion, which remains a significant challenge in video surveillance. \myNum{ii} \textit{MAVREC's color distribution} reflects European demographics, which may influence object detection algorithms that incorporate scene-contextual information, particularly those pre-trained on general object detection datasets; see a comparison in Figure \ref{fig:color_distribution}. \myNum{iii} Historically, {vehicle color distributions} vary across Europe, North America, and the Asia-Pacific; see Figure \ref{fig:car_color_dist_2012}.~The existing datasets collected in Asia and North America appear to be more colorful. E.g., in 2021, Europe's top car colors were gray (27\%), white (23\%), and black (22\%), contrasting with North America's gray (21\%), black (20\%), blue (10\%), red (~10-11\%), and silver (10\%), and China's predominance of white (50\%) and brown (10\%) cars \cite{car-color-1}.~\myNum{iv} MAVREC was collected at \textit{high latitudes}. The elevation of the sun in these areas (see Figure \ref{fig:solar_zenith}) during the peak traffic times is high, creating a {\em mirage-like} reflection on one of the sensors in many scenes, thereby, causing significant disparities between the two views. The second column of rows 1 and 2 in Figure \ref{fig:Different_scenes_from_the_dataset} shows this effect.   
\myNum{v} The aerial perspective inherent in MAVREC leads to \textit{small objects} inclusion; their presence is susceptible to miss-detection by detection algorithms. 
\myNum{vi} MAVREC is characterized by both \textit{sparse} and \textit{dense} distribution of objects. Our empirical findings suggest that such a large disparity in object distribution presents challenges in training object detectors, compared to scenes exhibiting only uniformly dense annotations.

\begin{table*}[t]
    \centering
    \small
    \small 
    \caption{{Supervised benchmark of MAVREC. D-DETR* denotes a MSCOCO pre-trained D-DETR.}}
    \setlength{\tabcolsep}{4.2pt}
    \begin{tabular}{ccccccccc cccccccc}
    \midrule
    \multirow{1}{*}{\textbf{Trained}} & \multicolumn{8}{c}{\textbf{Validation Set}} & \multicolumn{8}{c}{\textbf{Test Set}}\\
    \cmidrule(rl){2-9} \cmidrule(rl){10-17}
    \multirow{1}{*}{ \textbf{DNN}} & \multicolumn{4}{c}{\textbf{Ground}} & \multicolumn{4}{c}{\textbf{Aerial}} & \multicolumn{4}{c}{\textbf{Ground}} & \multicolumn{4}{c}{\textbf{Aerial}} \\
    \cmidrule(rl){2-5} \cmidrule(rl){6-9} \cmidrule(rl){10-13} \cmidrule(rl){14-17}
    \textbf{Models} & \texttt{AP} & ${\tt AP}_{50}$ &  ${\tt AP}_{\tt S}$ &  ${\tt AP}_{\tt M}$ & \texttt{AP} & ${\tt AP}_{50}$ &  ${\tt AP}_{\tt S}$ &  ${\tt AP}_{\tt M}$ &  \texttt{AP} & ${\tt AP}_{50}$ &  ${\tt AP}_{\tt S}$ &  ${\tt AP}_{\tt M}$ & \texttt{AP} & ${\tt AP}_{50}$ &  ${\tt AP}_{\tt S}$ &  ${\tt AP}_{\tt M}$\\
    \midrule
    DETR   & 21.8 & 36.9 & 21.9 & 23.9  & 24.9  & 39.7 & 27.6 & 45.3 & 20.8  & 35.4 & 21.3 & 24.0 & 23.6 & 40.1 & 23.4 & 44.9\\
    D-DETR & 27.5 & 51.4 & 28.1 & 43.7  & 13.1 & 28.3 & 14.2 & 38.1 & 18.2 & 46.8 & 17.9 & 36.0 & 10.3 & 25.0 & 10.1 & 29.4 \\
    {D-DETR}* & \textbf{59.6} & \textbf{82.7}  & \textbf{59.7} & \textbf{79.6} & 31.0 & \textbf{61.7} & 31.7 & 55.1 & \textbf{58.6} & \textbf{81.4} & \textbf{59.0} & \textbf{80.2} & \textbf{33.2} & \textbf{61.9} & \textbf{31.5} & 51.0 \\
    \midrule
    Yolo-NAS (L) & 41.4 & 61.7 & 36.8 & 72.9 & 30.3 & 49.8 & 29.2 & 61.5 & 41.2 & 63.4 & 37.8 & 74.3 & 27.0 & 43.3 & 25.9 & 58.0\\
    YoloV7 & 45.6 & 72.1 & 40.6 & 74.9 & \textbf{31.3} & 57.7 & \textbf{34.2} & \textbf{61.2} & 45.0 & 72.5 & 42.4 & 74.4 & 31.9 & 58.8 & 31.4 & \textbf{63.1} \\
    \midrule
    \end{tabular}\label{tab:baselines:supervised_benchmark}
\end{table*}

\vspace{-0.5ex}
\section{Baselines and evaluation}\label{sec:evaluation}
This section presents the benchmarking results on MAVREC in supervised and semi-supervised settings. We also present our observations concerning the prevailing trends in object detectors employed on aerial images.

\smartparagraph{Datasets and evaluation metric.} For supervised and semi-supervised benchmarking with \textbf{MAVREC}, we use a total of 8,605 labeled frames, and at most 8,605 unlabeled frames from each view at training. The validation and test set ({extracted from disjoint video sequences}) for each view contain 805 and 1,614 annotated images, respectively. 
We evaluate the models with the widely used metric for object detection, mean average precision (mAP) \cite{COCO}; {see a detailed discussion in \S\ref{app:metric}}. 

\smartparagraph{Object detector baselines.} For \textit{supervised benchmarking}, we use CNN-based YoloV7 \cite{yolov7}, and transformer-based DETR \cite{DETR} and D-DETR \cite{D-DETR}.~Additionally, we use Yolo-NAS \cite{yolonas}.~For \textit{semi-supervised benchmarking}, we propose a curriculum based semi-supervised baseline using D-DETR.
We provide the implementation details and computing environment in the \S \ref{sec:appendix_implementation}; we refer to Tables \ref{table:DNN_models} and \ref{table:experiments_hyperparameters} for other model specific implementation details. 

\subsection{Supervised benchmarking}\label{sec:supervised-benchmark}
Table \ref{tab:baselines:supervised_benchmark} presents the supervised baselines results on MAVREC dataset for both ground and aerial perspectives. Despite an equal number of training samples from different views, we observe that all the baselines exhibit superior performance on the ground perspective compared to the aerial perspective. This discrepancy highlights the challenge associated with object detection in aerial views due to their smaller sizes, as indicated by the ${\tt AP}_{\tt S}$ metric. Notably, YoloV7 demonstrates the best performance on aerial images, while D-DETR pre-trained on MSCOCO surpasses other models on the ground view. 
Interestingly, Yolo-NAS, which surpasses other Yolo-based detectors on ground images according to \cite{yolonas}, exhibits lower performance than YoloV7 on aerial images indicating that the learned Yolo-NAS architecture is suboptimal for aerial images. We show some qualitative results in Figure \ref{Figure:qualitative_results}.

\begin{figure*}
	\begin{minipage}[t]{0.45\textwidth}
		\centering
		\vspace{-23ex}	
		\resizebox{1.1\textwidth}{!}
		{
			\begin{tabular}{cccccc}
    \midrule
    \multirow{2}{*}{\textbf{Training Protocol}} & \multirow{2}{*}{\textbf{Pre-training}}  &  \multicolumn{4}{c}{\textbf{Test Set}} \\
    & & {\tt AP} & ${\tt AP}_{50}$  & ${\tt AP}_{S}$ & ${\tt AP}_{M}$  \\
    \midrule
    Trained from scratch   & \ding{53} & 10.3 & 25.0 & 10.1 & 29.4 \\
    Grounding-DINO & O365, GoldG, Cap4M & 20.4 & 40.9 & 18.6 & 32.5 \\
    FT on MAVREC Aerial view & Visdrone~\cite{visdrone}   & 20.9 & 41.9 & 20.6 & 43.8 \\
    FT on MAVREC Aerial view & MS-COCO~\cite{COCO} & 33.2 & 61.9 & 31.5 & 51.0 \\
    \rowcolor{gray!30} FT on MAVREC Aerial view & MAVREC Ground view   & \textbf{44.8} & \textbf{71.5} & \textbf{42.9} & \textbf{72.4} \\
    \midrule
    \end{tabular}}
		\vspace{-2ex}
		\captionof{table}{{Object detection using D-DETR~\cite{D-DETR} on the aerial view images of the proposed MAVREC dataset; FT indicates finetuning.}}
		\label{tab:intro_table}
	\end{minipage}\hfill
	\begin{minipage}[b]{0.5\textwidth}
\includegraphics[width=\columnwidth,height= 1.2in]{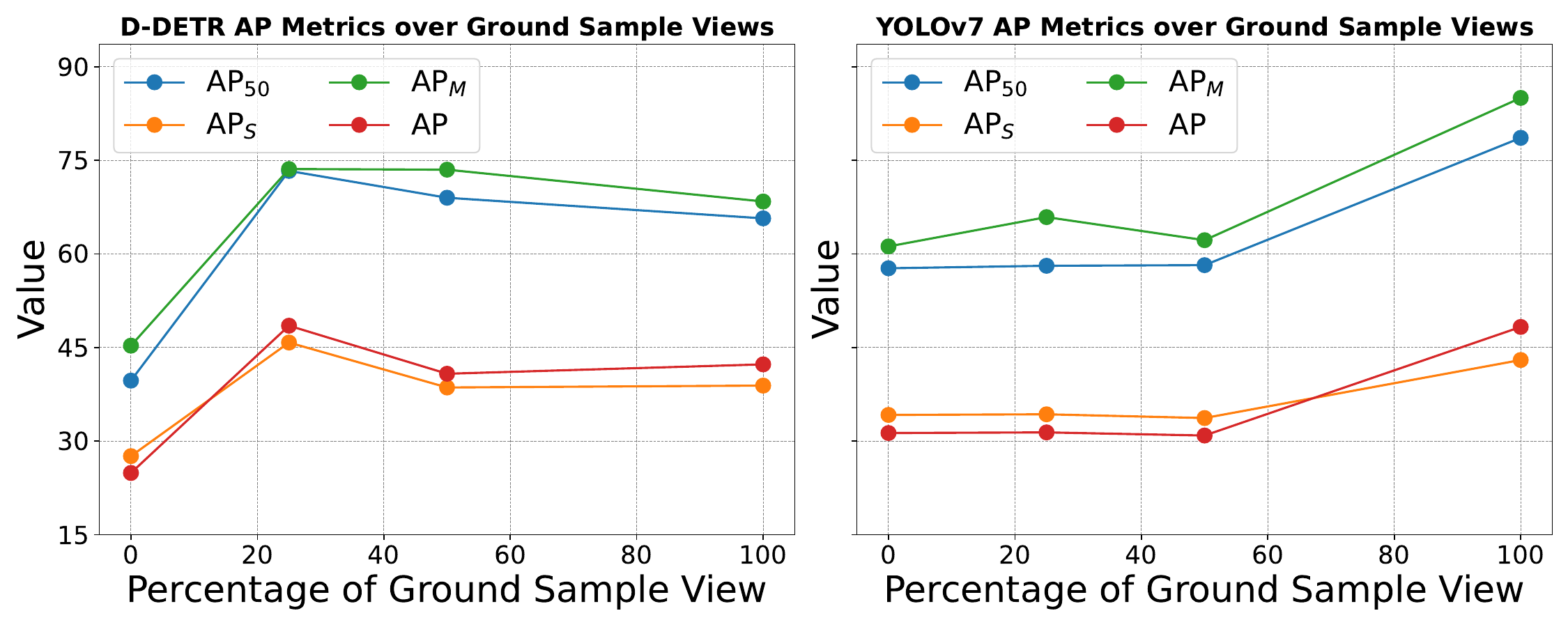}
\captionof{figure}{{Supervised benchmark on aerial view of MAVREC (Validation Set).}}
		\label{fig:mavrec_benchmark}
	\end{minipage}\hfill
	\caption*{}
\vspace{-3ex}
 \end{figure*}

\begin{figure*}[ht]
\centering
\includegraphics[width=.19\textwidth]{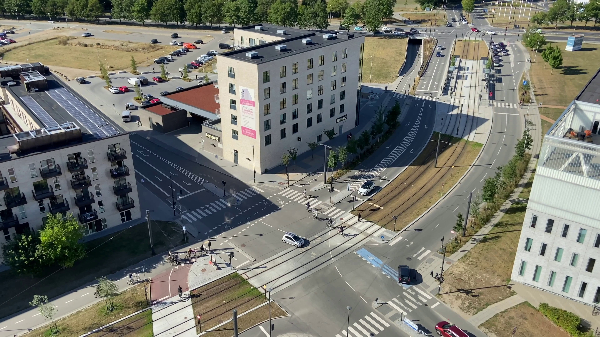}
\hfill
\includegraphics[width=.19\textwidth]{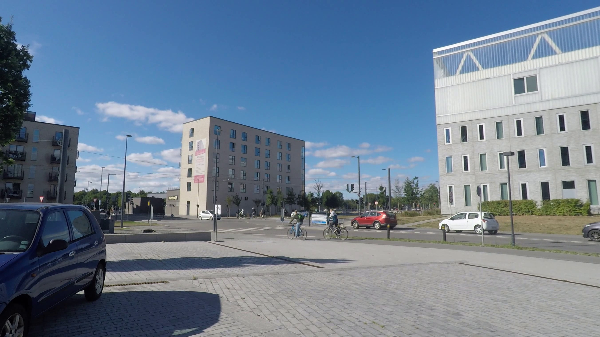}
\hfill
\includegraphics[width=.19\textwidth]{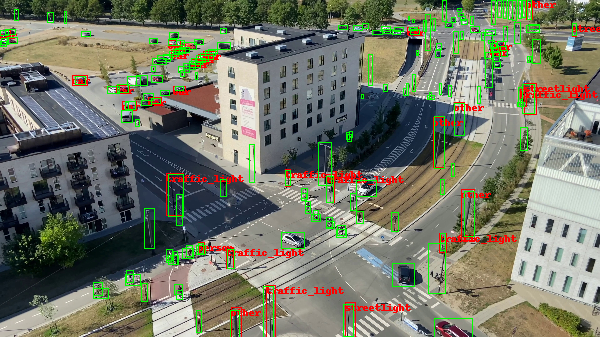}
\hfill
\includegraphics[width=.19\textwidth]{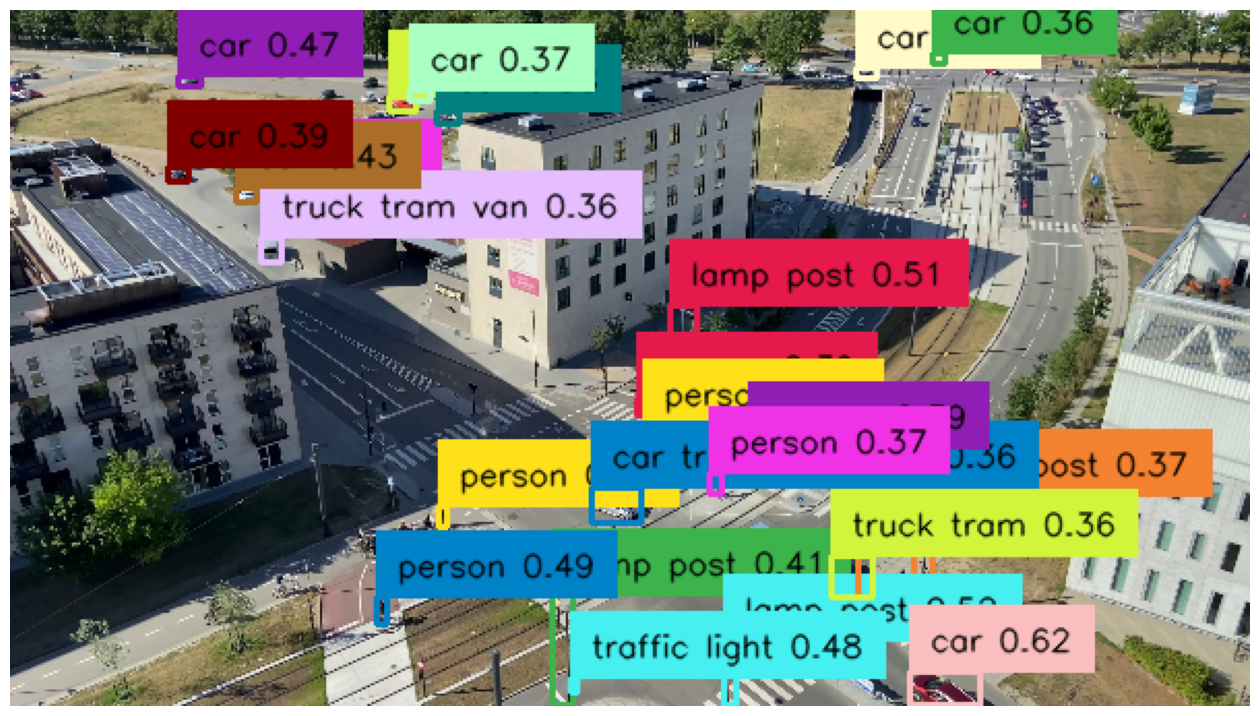}
\hfill
\includegraphics[width=.19\textwidth]{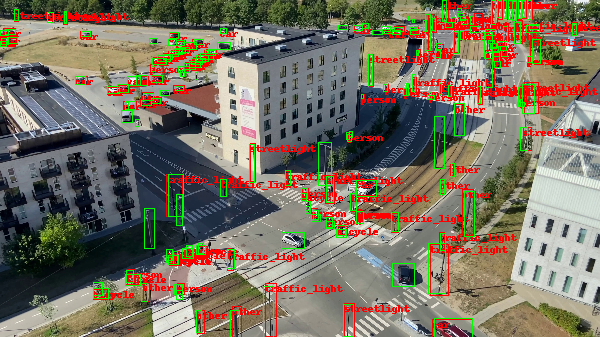}
\caption{\textbf{Left to right}: Sample frames from time-synchronized MAVREC showing aerial and ground views; D-DETR~\cite{D-DETR} trained on aerial VisDrone DET \cite{visdrone} inference results on MAVREC~(GT bounding boxes are \textcolor{green}{green}, and detection are in \textcolor{red}{red}); Grouding-Dino \cite{kirillov2023segany} inference result on MAVERC; inference results of D-DETR trained on aerial MAVREC has fewer missed detections. (Zoom in for a better view)}
\label{fig:teaser}
\end{figure*}
\subsection{Curriculum learning based semi-supervised object detection}\label{sec:semi-supervised-benchmark}

\smartparagraph{Can ground-view images improve object detection in aerial perspective?}
To answer this, we trained D-DETR and YoloV7 by augmenting the existing aerial-view sample set with ground-view samples.~We achieve this by simply adding the two sets of aerial- and ground-view samples along with their corresponding annotations. Our findings demonstrate that the inclusion of ground-view samples substantially improves the object detection performance.

Graphs presented in Figure~\ref{fig:mavrec_benchmark} illustrates that D-DETR outperforms the CNN-based YoloV7 when the extra ground-view samples enrich the training distribution. While YoloV7 requires an equal number of ground-view samples as aerial samples to achieve its peak performance, D-DETR achieves a relative improvement upto 270\% with a subset of ground-view samples ($\sim2$K ground-view images). Interestingly, further augmentation of ground-view images during D-DETR training does not enhance its performance, indicating the sensitivity of D-DETR's training process to ground-view image sampling. 

Thus, experimental analysis of MAVREC within a supervised framework suggest that 
\myNum{i} CNN-based models, such as Yolo, demonstrated superior performance over transformer-based models, like D-DETR, when trained from scratch. However, transformer based model outperforms when pretrained on large-scale ground images in MSCOCO. This shows the superiority of these architectures when large-scale data is available. \myNum{ii} These transformer based object detectors augmented with even 25\% of MAVREC's ground view images, surpasses the model pretrained on MSCOCO. This shows the importance of learning geography-aware representation for aerial visual perception, suggesting a new direction for enhancing object detection in aerial perspective. 

\begin{figure}[t]
\centering
\scalebox{0.9}{
    \includegraphics[width=\columnwidth]{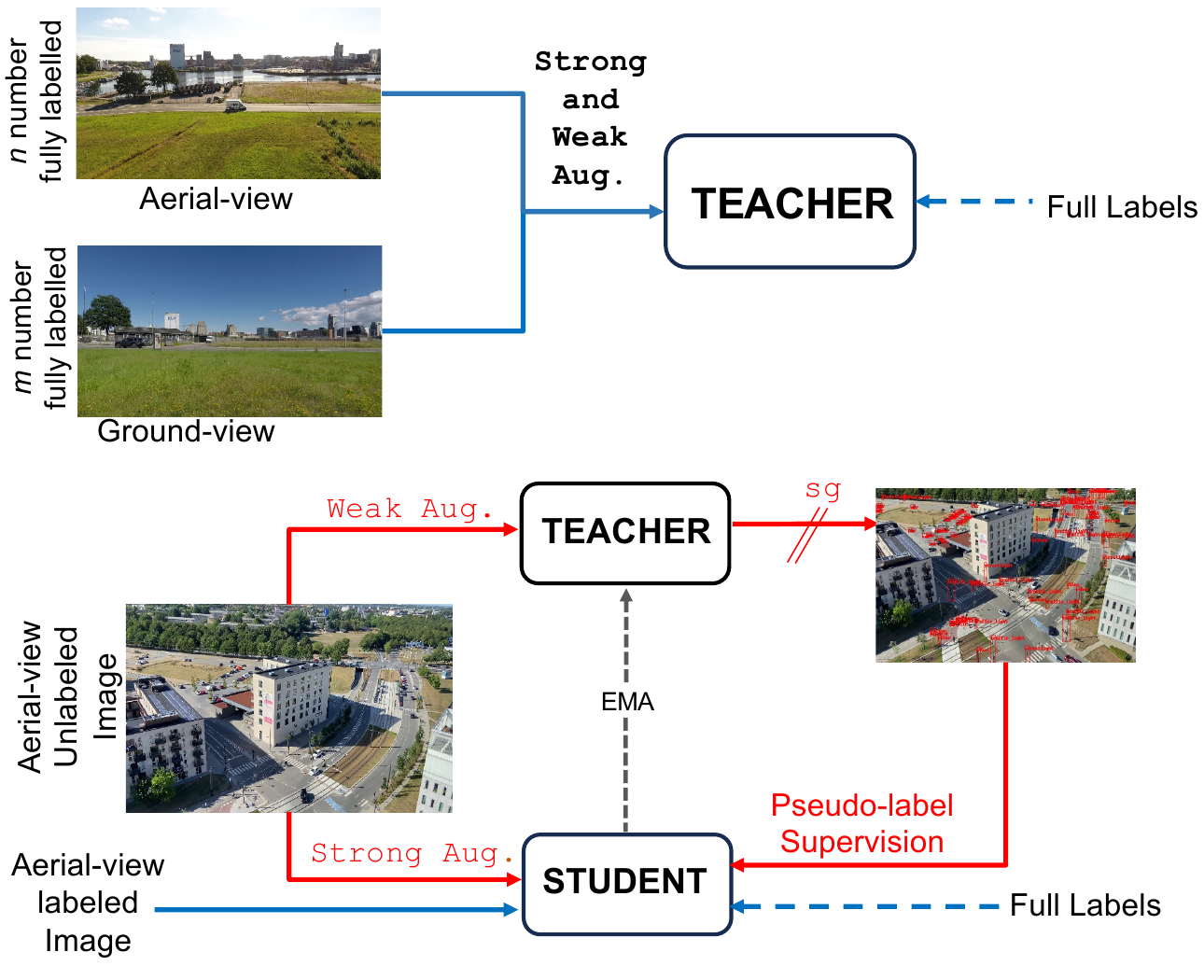}}
            \caption{{Semi-Supervised object detection framework based on curriculum learning approach. Here \textcolor{blue}{blue} represents the initial supervised stage, \textcolor{red}{red} represents the later unsupervised stage, SG represents stop gradient.}}
            \label{fig:Curriculum Learning}
\vspace{-1ex}
\end{figure}
\noindent \textbf{Model generalization.} 
In Table~\ref{tab:intro_table}, we provide the object detection performance of MAVREC with models pretrained with different strategies. 
Our empirical evaluations indicate that state-of-the-art object detection models,including the \emph{open-world foundational models} like Grounding-Dino~\cite{liu2023grounding}, which fail to achieve the expected performance level on MAVREC; see Figure~\ref{fig:teaser}.~This observation validates an inherent bias of these models towards ground-view data.~Moreover, Table~\ref{tab:intro_table}, combined with the visual evidence in Figure \ref{fig:teaser}, shows an object detector pre-trained on popular ground-view dataset (MS-COCO~\cite{COCO}) or other aerial datasets collected from different geographies (e.g., Visdrone~\cite{visdrone} from China) has diminished efficacy on aerial images obtained from disparate geographical regions (for our case, Europe). Therefore, unlike classical object detection, training a sophisticated DNN model on a large dataset (e.g., ImageNet \cite{ImageNET} or MS-COCO \cite{COCO}) does not offer the best overall solution. We find augmenting object detectors with ground-view images from the corresponding geographical context is a superior strategy that boosts detection performance.

\begin{table*}
    \centering
    \small
\footnotesize
\caption{{Semi-supervised Omni-DETR \cite{OMNI-DETR} benchmark on MAVREC. In the table, $G$ and $A$ denote number of ground and aerial-view images, respectively. During the burn-in, we only use the labelled subset.}}
    \begin{tabular}{c cccc cccc ccccc}
    \midrule
    \textbf{Training} & \multicolumn{2}{c}{\textbf{Labelled}} & \multicolumn{2}{c}{\textbf{Unlabelled}} & \textbf{Test}& \multicolumn{4}{c}{\textbf{Validation Set}} & \multicolumn{4}{c}{\textbf{Test Set}}\\
    \cmidrule(rl){2-3}  \cmidrule(rl){4-5} \cmidrule(rl){7-10} \cmidrule(rl){11-14}
    \textbf{Technique} & ${G(m)}$ & ${A(n)}$ & ${G}$ & ${A}$ & \textbf{perspective} & \texttt{AP} & ${\tt AP}_{50}$ &  ${\tt AP}_{\tt S}$ &  ${\tt AP}_{\tt M}$ & \texttt{AP} & ${\tt AP}_{50}$ &  ${\tt AP}_{\tt S}$ &  ${\tt AP}_{\tt M}$\\
    \midrule
    Semi-Supervised & 8605 & 0 & 8605 & 0  & G  & 56.9 & 83.3 & 54.8 & 74.9  & 45.8 & 75.5 & 45.4 & 58.7 \\ 
    \midrule
    Semi-Supervised & 0 & 8605 & 0 & 8605  & A    & 29.3 & 49.3 & 24.8 & 60.6 & 19.8 & 38.4 & 19.5 & 35.0 \\
    \midrule
    \midrule
    \rowcolor{gray!30}
    \textbf{\textit{Curriculum Learning (Ours)}} & 2151 & 8605 & 0 & 8605 & A    & 37.8 & \textbf{64.9} & 35.4 & 68.3 & 23.2 & 45.4 & 21.8 & 43.6 \\
    \midrule
    \rowcolor{gray!30}
    \textbf{\textit{Curriculum Learning (Ours)}} & 2151 & 8605 & 2151 & 8605 & A  & \textbf{38.0} & 64.8 & \textbf{35.7} & 67.6  & \textbf{26.7} & \textbf{54.1} & \textbf{24.5} & 42.4\\
    \midrule
    \end{tabular}
    \label{tab:baselines:semi_supervised_benchmark}
\end{table*}

In this section, we introduce a curriculum based learning strategy for semi-supervised object detection. Curriculum learning \cite{soviany2022curriculum} provides a systematic strategy to enhance model performance by incrementally introducing complexity into the training regime. We adopt curriculum learning in the semi-supervised object detection using a D-DETR.
In our semi-supervised baseline, we train the object detector using both labeled and unlabeled image sets. The foundation of our semi-supervised baseline is a consistency regularization framework based on a {\tt teacher-student} model. This framework encompasses two distinct training phases: \myNum{i} the `burn-in' stage, where the {\tt teacher} model is trained exclusively with labeled images, and \myNum{ii} the semi-supervised training stage, where the {\tt teacher-student} model engages with unlabeled images.
Given a trained {\tt teacher} network from the burn-in stage, this framework leverages weak-to-strong consistency regularization \cite{fan2023revisiting} to leverage unlabeled aerial images as shown in Figure~\ref{fig:Curriculum Learning}.  
In this stage, the {\tt teacher} network processes the weakly augmented unlabeled aerial-view images to generate bounding boxes. These bounding boxes serve as pseudo-labels for the strongly augmented counterpart of the image, which is the input of the {\tt student.} The {\tt teacher} network is updated through an exponential moving average (EMA) of the {\tt student}'s updates, and the {\tt student} network only gets back-propagated.

However, the performance of the above semi-supervised baseline partially relies on the effectiveness of the {\tt teacher} network to generate pseudo-labels. Inspired from our insights in transferring geography-aware knowledge from ground view to aerial view, we employ a curriculum learning strategy in the burn-in stage of the semi-supervised framework. In contrast to training the {\tt teacher} network with only labelled aerial images, we train the teacher, first, with $m$ labelled ground-view images and then with $n$ labelled aerial images. The outcome is a trained geography-aware {\tt teacher} network that facilitates the second phase of training the semi-supervised framework by generating precise object proposals.

In Table~\ref{tab:baselines:semi_supervised_benchmark}, we showcase the results of semi-supervised benchmarking on the MAVREC dataset, utilizing D-DETR as the backbone object detection model within the teacher-student framework. Note that all the models are trained on 39 epochs with 20 epochs for the burn-in stage and 19 epochs for the semi-supervised training stage.
Our semi-supervised baseline results demonstrate that by utilizing the same number of unlabeled aerial images as labeled images, we achieve a substantial boost in object detection performance---from 13.1\% to 29.3\% and 10.3\% to 19.8\% on validation and test set, respectively. We observe a consistent improvement in the ground view. This warrants the importance of using semi-supervised approaches for vision tasks, particularly where the annotation process is labor-intensive.
Subsequently, we demonstrate the performance of our curriculum-based semi-supervised object detector on the MAVREC dataset. This approach outperforms the baseline model by 8.5\% and 7\% on validation and test set, respectively. Moreover, we observed additional improvements in object detection accuracy when the second phase of training is augmented with unlabelled ground-view images. This shows the importance of learning geography-aware representation from the ground-view for enhancing aerial visual perception.

\section{Conclusion}\label{sec:conclusion}
In this paper, we introduce a large-scale, high-definition ground and aerial-view video dataset, MAVREC. To the best of our knowledge, MAVREC is the first drone-based aerial object detection dataset that exploits the multi-view of the data coming from orthogonal views, aerial and ground to offer enhanced detection capacity for an aerial view. 
In our extensive benchmarking on MAVREC, we employed both supervised and semi-supervised learning methods, along with our proposed curriculum-based ground-view pre-training strategy. Our findings highlight several key insights, particularly the importance of geographic awareness in aerial visual models. We discovered that models trained on aerial images from one geographic location often struggle to generalize to different regions. However, integrating ground-view data from the same geographic area significantly enhances the model's ability to learn more distinctive visual representations. 
We envision that this dataset and benchmarking will benefit: \myNum{i} researchers, who will use it as the basis for consistent implementation and evaluation; and \myNum{ii} practitioners, who need an appropriate, large-scale, industry-standard dataset for training DNN models for aerial images. 


{
    \bibliographystyle{ieeenat_fullname}
    \bibliography{egbib}
}
\clearpage

\appendix
\noindent \textbf{\huge Appendix}
\begin{figure*}[t]\mbox{\includegraphics[width=\textwidth]{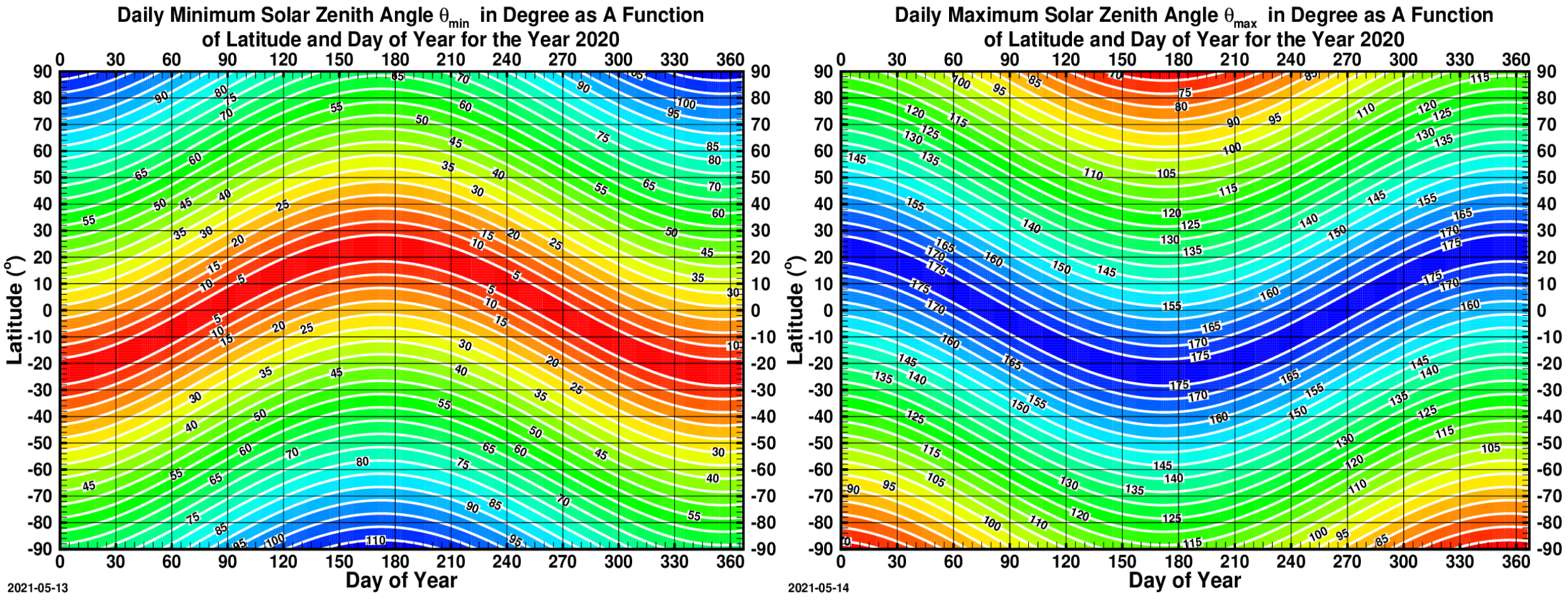}}		
			\caption{{The daily minimum and maximum of the solar zenith angle as a function of latitude and day of year for the year 2020. In the Earth-Centered Earth-Fixed (ECEF) geocentric Cartesian coordinate system, let $(\phi_{s}, \lambda_{s})$ and $(\phi_{o}, \lambda_{o})$ be the latitudes and longitudes of the subsolar point and the observer's point, then the upward-pointing unit vectors at the two points, $\mathbf{S}$ and $\mathbf{V}_{oz}$, are $\mathbf{S}=\cos\phi_{s}\cos\lambda_{s}{\mathbf i}+\cos\phi_{s}\sin\lambda_{s}{\mathbf j}+\sin\phi_{s}{\mathbf k},$ and $\mathbf{V}_{oz}=\cos\phi_{o}\cos\lambda_{o}{\mathbf i}+\cos\phi_{o}\sin\lambda_{o}{\mathbf j}+\sin\phi_{o}{\mathbf k},$ where ${\mathbf i},{\mathbf j}$ and ${\mathbf k}$ are the basis vectors in the ECEF coordinate system. Consequently, cosine of the solar zenith angle, $\theta_{s}$, is the inner product between $\mathbf{S}$ and $\mathbf{V}_{oz}$.~Source: \cite{wiki-zeneith}.}}
    \label{fig:solar_zenith}
 \end{figure*}
\section{Related work---Continued}\label{app:relatedwork}
This section extends the discussion in Section \ref{sec:related_work} of the main paper by including additional UAV-based datasets that focus on different downstream tasks such as action detection, counting, geo-localization, 3D reconstruction, and benchmarking; also, see \cite{wu2022_survey}.

\myNum{i}\smartparagraph{Human, vehicle, and drone trajectory tracking.}~{PNNL~1 and~2}~\cite{PNNL} {are unannotated datasets consisting of 1,000 and 1,500 frames, respectively, designed for human tracking from a fixed perspective with long-term inter-object occlusion.~{The highway-drone dataset~\cite{highDdataset} is a large-scale dataset collected from 6 different locations on German highways, crafted for the safety validation of automated vehicles. The dataset consists of more than 110,500 vehicle annotations, recorded over 147 hours, and offers each vehicle's trajectory, including type, size, and maneuvers.} Among others, \textit{UVSD} \cite{UVSD} is a small-scale (5,874 images), multi-view, aerial dataset for vehicle detection and segmentation. \textit{DroneVehicle} \cite{DroneVehicle} (thermal infra-red+RGB) and \textit{BIRDSAI} \cite{BIRDSAI} (thermal infra-red) are small-scale, low-resolution datasets used for detection, tracking, and counting. 

\textit{MVDTD}~\cite{MVDTD} is a collection of datasets to estimate 3D drone trajectories from multiple unsynchronized cameras. \textit{UAVSwarm}~\cite{UAVSwarm} detects and tracks UAVs.
~\cite{MTDTUAVs} proposes drone-to-drone detection and tracking from a single drone-camera.~\textit{EyeTrackUAV2}~\cite{ETUAV2} tracks drones from a ground perspective, specifically, from a {\em binocular} viewpoint.

\myNum{ii}\smartparagraph{Action detection from aerial viewpoints.}~{UCF-ARG} \cite{UCF_ARG} is a multi-view, scripted dataset, designed for 10 different human action detection, where the scenes are recorded from 3 different views---a rooftop camera, a ground camera, and an aerial camera.~{Okutama-Action}~\cite{Okutama-Action} is an aerial dataset consisting of 77,365 annotated frames, designed for 12 concurrent human action detection. 

\myNum{iii}\smartparagraph{Counting and 3D reconstruction.}
{CARPK}~\cite{CARPK} is a single-view video dataset, captured from a moving drone, contains nearly 90,000 cars from 4 different parking lots, and is used for predicting the car-counts in a scene.~{CarFusion}~\cite{CarFusion} is a multi-view dataset consisting of 53,000 fully-annotated frames, 100,000 car instances with 14 semantic key points, captured from 18 moving cameras at multiple locations, designed for 3D reconstruction of cars. 

\myNum{iv}\smartparagraph{Geo-localization} is a challenging problem, and over the past years, some dedicated datasets were proposed to devise efficient solutions to this problem. Danish airs and grounds (DAG) dataset \cite{vallone2022danish}  is a large collection of ground-level and aerial images covering about 50 kilometers in urban and rural environments with the extreme viewing-angle difference between query and reference images is a dataset for place recognition and visual localization. Similar to DAG, \cite{geo-localization3} assembled a much smaller dataset with a drone and GoogleMap images. For more details in this context, refer to \cite{geo-localization1,geo-localization2}.

\myNum{v}\smartparagraph{Other downstreaming tasks.}~{SeaDronesSee}~\cite{SeaDronesSee} is curated for single and multi-object tracking, specifically people, floating in water.~{DroneSURF} \cite{DroneSURF} is for person identification, especially facial recognition, in an urban environment, while \cite{DTCDrones} works on object detection, tracking, and counting.~{P-DESTRE} \cite{P-DESTRE} is a dataset designed to test pedestrian detection, tracking, re-identification, and search methods.~{VIRAT} \cite{VIRAT} is a video dataset from surveillance cameras, designed for testing on real-world environments and challenges. 

\myNum{vi}\smartparagraph{Benchmarking and evaluation.}~The {UAV Benchmark}~\cite{UAVBench} and \cite{VOTUAV} present datasets that maximize their breadth of usability, and provide extensive comparisons, including camera motion estimation. Finally, in \cite{wu2022_survey}, Wu et al.\ provides challenges and statistics of existing DL based methods for UAV-based object detection and tracking. 


\begin{figure*}
	\begin{minipage}[t]{0.5\textwidth}
		\centering
		\vspace{-40ex}	
		\resizebox{1.1\textwidth}{!}
		{
			\begin{tabular}{lll}
\midrule
Drone/UAV & DJI Phantom 4, DJI mini 2                          \\ 
ISO Range & 100-3200\\
Lens & FOV $94^{\circ}$ 20 mm, FOV $83^{\circ}$ 20 mm\\
\midrule 
GoPro & GoPro HERO4, HERO 6                             \\ 
ISO range & 100-800\\
\midrule 
iphone & 11, 13-Pro (when UAV not used)   \\ 
FOV & $120^{\circ}$\\
\midrule 
Resolution (GoPro, Drone) & 2.7K (2704x1520) \@ 30fps  \\ 
Filetype video & .mp4 (.mov)                     \\
Filetype image & {.png}                          \\ 
\midrule
\end{tabular}}
		\captionof{table}{\small{Details of the recording devices.}}
		\label{tab:hardware_specification}
	\end{minipage}\hfill
	\begin{minipage}[b]{0.45\textwidth}
\includegraphics[width=\textwidth]{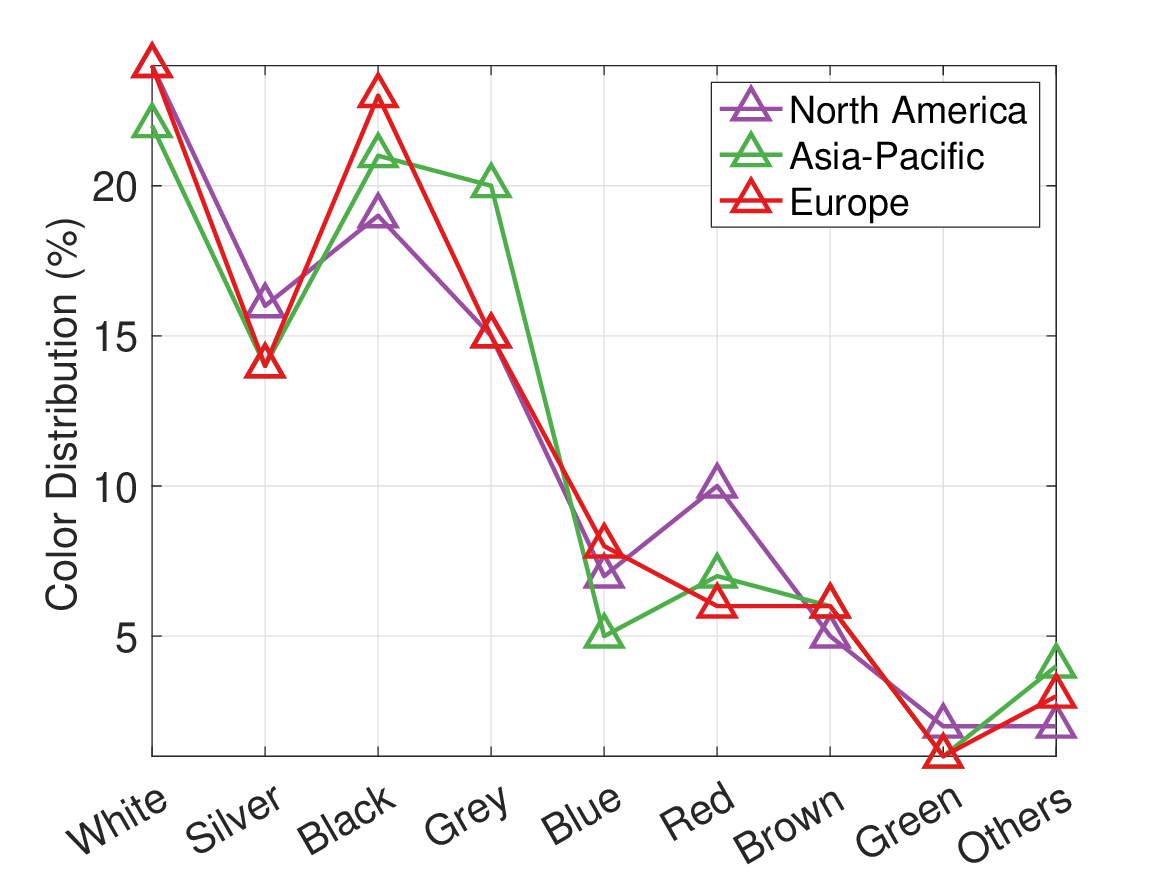}
\captionof{figure}{\small{Car color popularity surveys conducted by American paint manufacturer DuPont for the year 2012. Source: \cite{wiki-car}.}}
\label{fig:car_color_dist_2012}
\end{minipage}\hfill
	\caption*{}
 \end{figure*} 
 
 			
\begin{figure*}
    \centering
    \includegraphics[width=\textwidth]{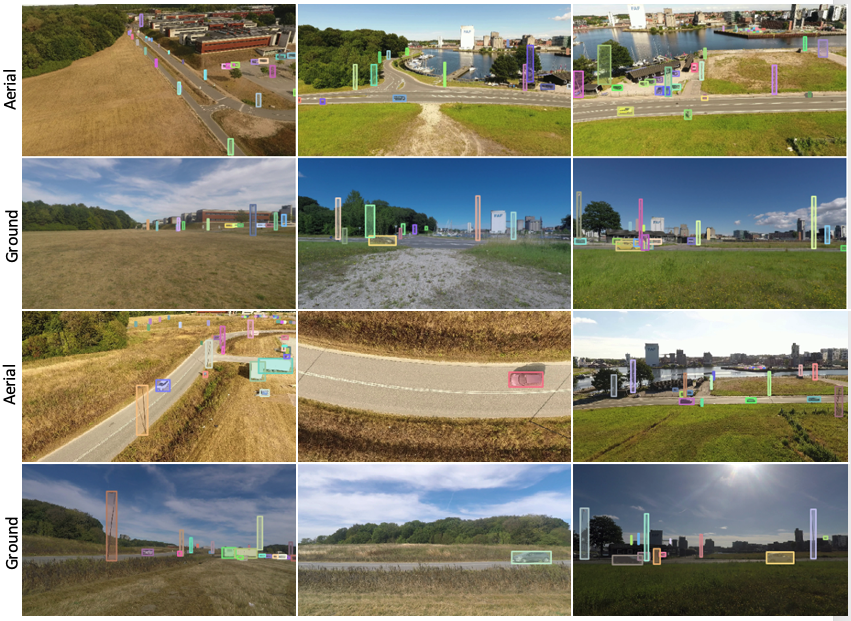}
    \caption{{Different sample scenes (with annotation) from our dataset; the first row is the aerial-view, second row presents the same scenes from a ground camera. Similarly, the third row is the aerial-view, and the fourth row presents the same scenes from a ground camera. Some scenes have a dense object annotations, while some scenes have very few object annotations. This high variance in object distribution across different scenes in MAVREC is complementary to datasets like VisDrone \cite{visdrone} where object detection is relatively straightforward due to their biased object distribution (dense), reflecting its demographic characteristics.}}
    \label{fig:Different_scenes_from_the_dataset_extra}
\end{figure*}

\begin{figure*}[t]
\centering
\includegraphics[width=0.9\textwidth]{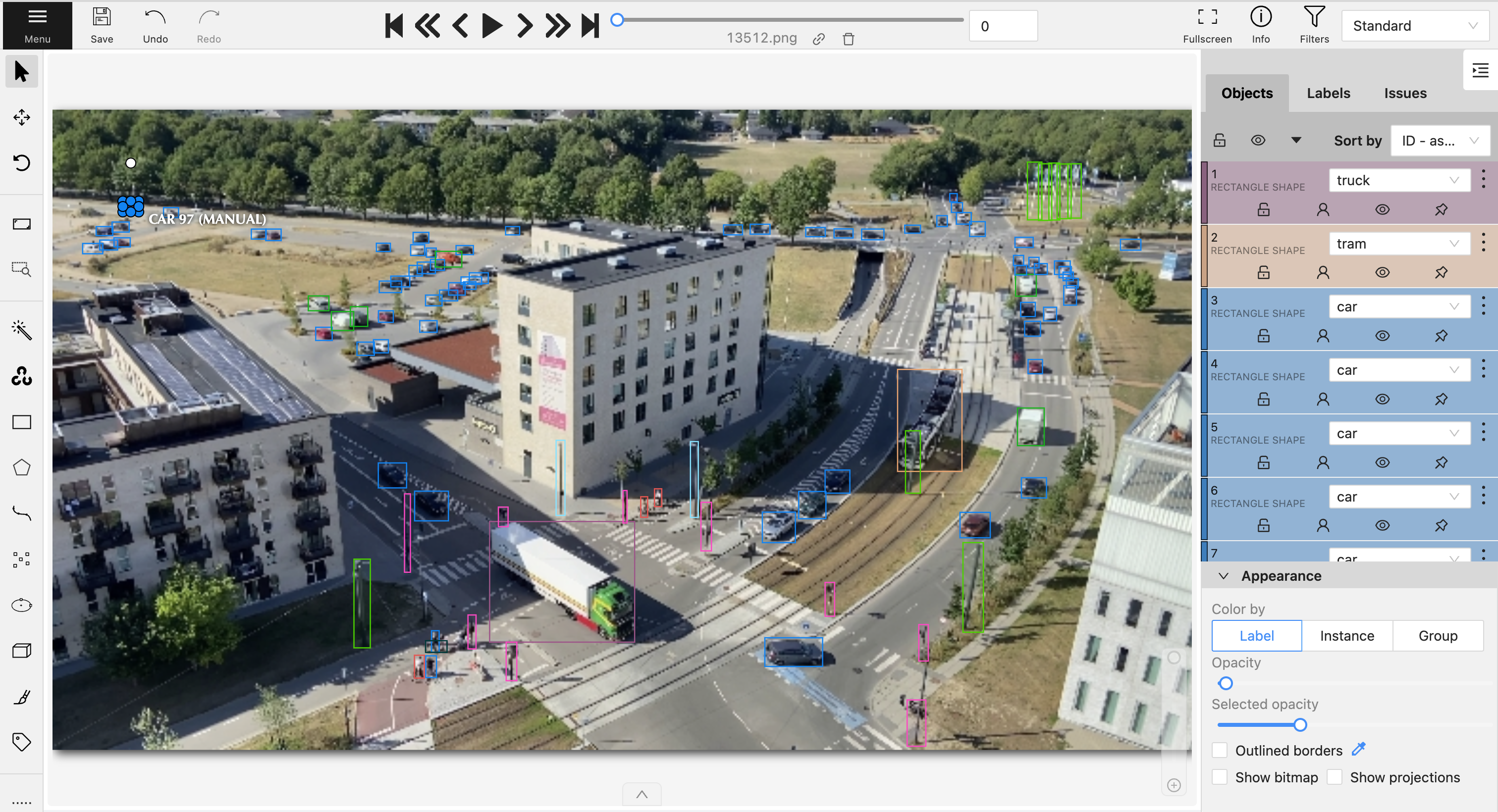}
\caption{\textbf{A sample annotation using {\tt CVAT} \cite{CVAT} interface.} {\tt CVAT} has an in-built tracker that tracks an object through multiple frames. The inbuilt tracker speeds up the annotation part --- once a particular frame is annotated, around 10 frames after that require minimal human supervision --- leveraging the tracker. This property makes {\tt CVAT} an attractive annotation tool.}\label{fig:CVAT}
 \end{figure*}
 
\section{Addendum to the dataset}\label{sec:Appendix-Dataset}
In this section, we provide some extra insights on the structuring and statistics of the MAVREC. Additionally, we discuss about the {\tt CVAT} annotation tool in Section \ref{sec:cvat}, and provide an analysis of color distribution of different drone based datasets and contrast them with MAVREC; see Section \ref{sec:Appendix-color}.
\begin{table*}
    
\caption{{Summary of annotations in both views of MAVREC.}}\label{tab:annotation_summary}
\vspace{-1ex}
\small
\begin{center}
\begin{tabular}{@{}lllcccc@{}}
 \multicolumn{5}{c}{} \\
 \midrule
\textbf{View} & \textbf{Train set} & \textbf{Test set}  & \textbf{Validation set} & \textbf{Total}  & \textbf{Total}  & \textbf{Annotations} \\
& \textbf{annotations} & \textbf{annotations}  & \textbf{annotations} & \textbf{annotations} & \textbf{annotated frames} & \textbf{per frame}  \\
\midrule
\textbf{Aerial} & 655,608 & 120,517 & 42,927 & 819,052 & 11,024 & 74.23 \\
 \midrule
\textbf{Ground} & 226,461 & 42,440 & 14,651 & 283,552 & 11,024 & 25.72\\
\midrule
\textbf{Combined} & 882,069 & 162,957 & 57,578 & 1,102,604 & 22,048 & 50.01\\
\midrule
\end{tabular}
\end{center}
\end{table*}
\begin{figure*}[t]
\centering
\includegraphics[width=0.65\textwidth]{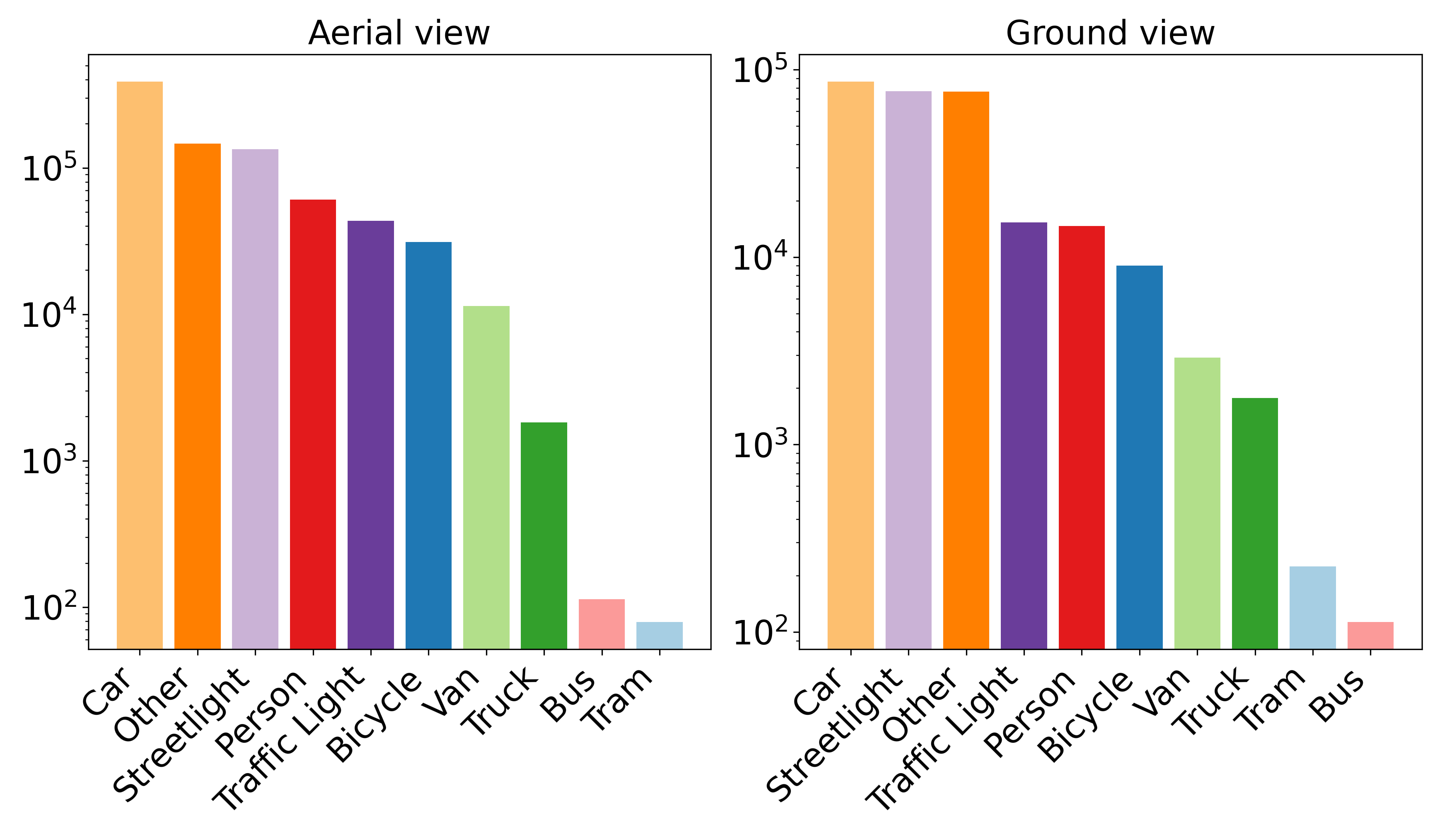}
\caption{Total numbers of objects in each category in the aerial and ground view.}
\label{fig:class_distribution-appendix}
\end{figure*}
\begin{figure*}[t]
     \centering
     \includegraphics[width=\textwidth]{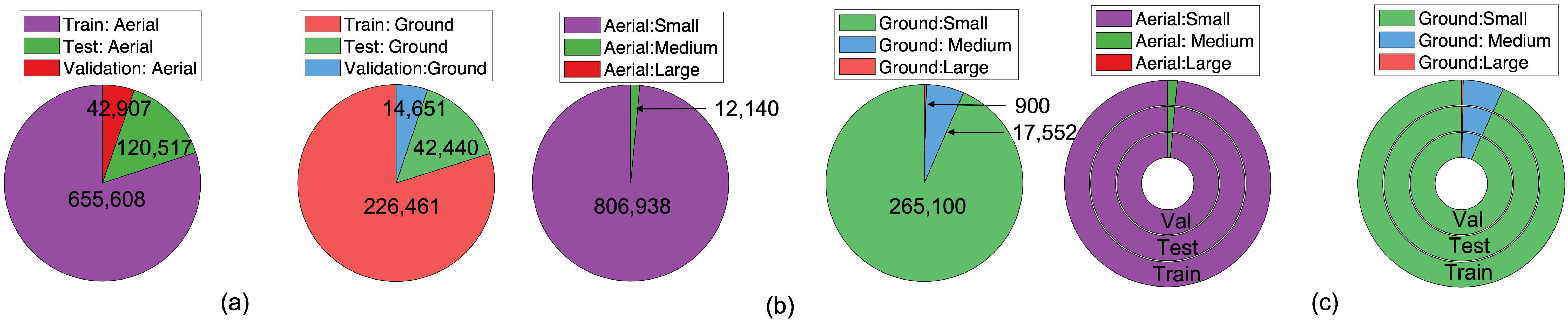}
     \caption{{(a) Total number of annotations in train, test, and validation sets of aerial and ground view; (b) number of objects based on their sizes in aerial and ground view, aerial view has no {\em large} object annotation; (c) percentage of small, medium, and large objects in train, test, and validation sets of aerial and ground view.}}
     \label{fig:object_size_distribution}
\end{figure*}


\subsection{{\tt CVAT} annotation tool}\label{sec:cvat}
{\tt CVAT} is an industry-standard, open-source, cutting-edge, interactive annotation tool that produces professional-level image and video annotations for diverse computer vision tasks \cite{CVAT}. {\tt CVAT} is equipped with an in-built tracker that can track an object consecutively for a few frames and results in an easier and faster annotation. Annotating in {\tt CVAT} is done by annotating category by category. This can either be done frame by frame or within an interval of frames relying on the built-in tracker for the frames in between. Figure \ref{fig:CVAT} presents one such instance of annotation interface using {\tt CVAT.}

\subsection{Color distributions of different datasets---An experimental analysis}\label{sec:Appendix-color}
The color content of different geographies on the earth is quite diverse. Many recent studies show that the latitude influences the solar elevation, and hence the population density \cite{kummu2011world, latitude} of different parts of the world. These factors have a direct effect on {\em color-content of the scenes}. In this scope, we analyze the color content of sample video frames from different datasets based on two key points: \myNum{i} color distribution in the sample frames of different datasets based on {\tt RGB} color channels, and \myNum{ii} dominant color distributions in the sample frames of the datasets.

\smartparagraph{Color distribution of different datasets based on {\tt RGB} color channels.} We show the color distributions of sample frames from different datasets in Figure \ref{fig:color_distribution}. For each dataset, we randomly sample 1000 images. All images are resized to $600 \times 337$ and an {\em average image} is computed. Then, a color histogram is computed for each color channel of the {\em average image}, and the area under each curve representing each color channel is calculated. Except for UAV123, the area under the green channel for all other datasets is about 1.5-2$\times$ lower than the MAVREC aerial view. However, the blue color channel of MAVREC is the most dominant in the aerial view. Additionally, the distribution of the blue and green channels in the ground view of the MAVREC are doubly-peaked, covering almost similar areas under them.  

\begin{figure*}[t]
     \centering
     \includegraphics[width=\textwidth]{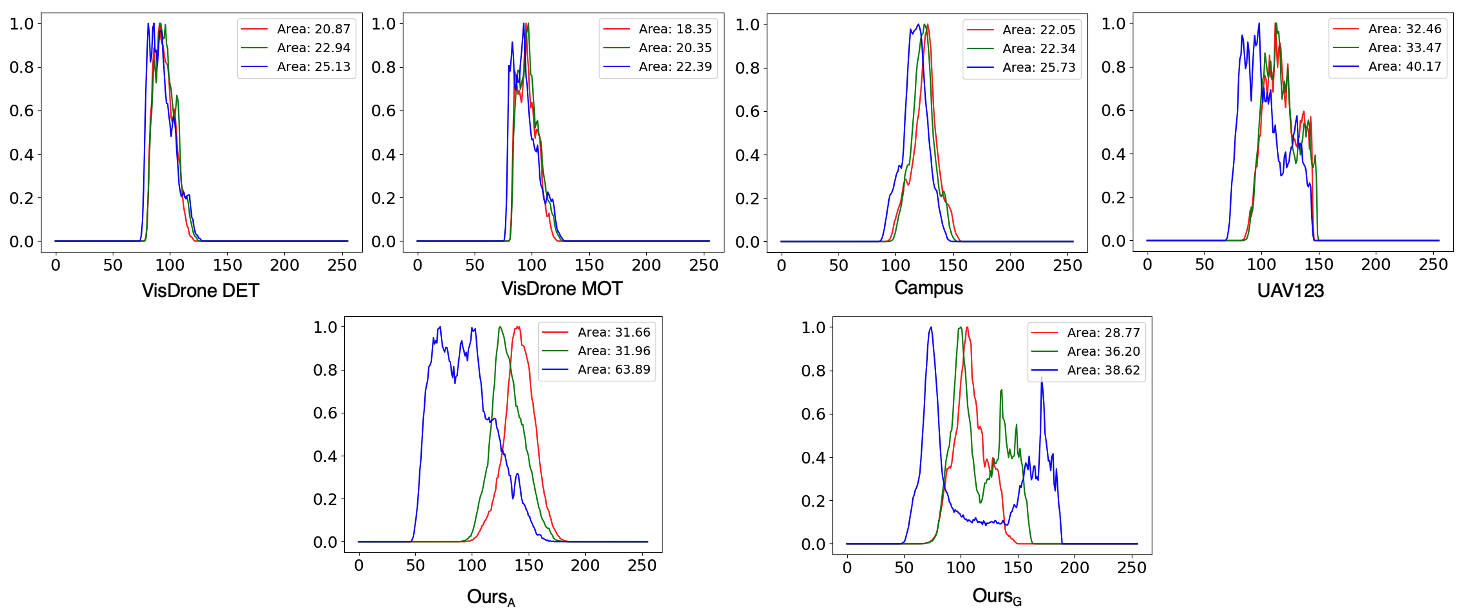}
     \caption{{\textbf{Color distribution of different datasets.} In the top row, we show the color distribution of VisDrone \cite{visdrone} DET and MOT, the Campus dataset \cite{campus}, and the UAV123 dataset \cite{UAVDT}. VisDrone represents south-east Asian geographies (collected in 14 cities across China) \cite{visdrone}; the Campus dataset represents North American geographies, collected in Stanford University campus \cite{campus}; UAV123 represents the Middle East, collected primarily in King Abdullah University of Science and Technology's campus and its surroundings (Kingdom of Saudi Arabia) \cite{UAV123}. In the bottom row, we show the ground and aerial view color distribution of MAVREC.}}
     \label{fig:color_distribution}
\end{figure*}

\smartparagraph{Dominant colors in MAVREC and other datasets.} 
 We use the {\tt Python} tool {\tt extract-colors-py}, which groups colors based on their visual similarities by using the {\tt CIE76} standard \cite{CIE76}.~The tool, {\tt extract-colors-py} uses two hyperparameters: \myNum{i} the tolerance, $\epsilon$, that determines how two colors can be grouped (default $\epsilon=32$), and \myNum{ii} color limit, that is the upper limit of extracted colors in the output. We set both the $\epsilon$ and the color limit to 12 and plot the grouped colors with their percentages. In Figure \ref{fig:color_dominans_in_MAVREC}, we analyze the most dominant colors in MAVREC in different sample scenes (aerial and ground), while Figure \ref{fig:color_dominans_in_comparing_datsets} shows the dominant colors in other datasets. Indeed, the dominance of different spectra of blue, yellow, and green colors in MAVREC in both views as shown in Figure \ref{fig:color_dominans_in_MAVREC} directly supports our findings in Figure \ref{fig:color_distribution}, and make MAVREC a stand-alone video dataset compared to the other large-scale, drone-based datasets such as VisDrone~\cite{visdrone}, UAV123~\cite{UAV123}, Campus~\cite{campus}. 
\begin{figure*}[t]
  \begin{minipage}[t]{0.48\textwidth}
    \begin{subfigure}{\linewidth}
          \begin{minipage}{0.48\linewidth}
            \includegraphics[width=\linewidth]{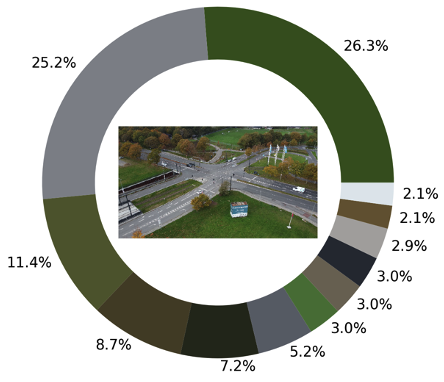}
            \caption{Scene 1 Aerial}
          \end{minipage}\hfill
          \begin{minipage}{0.48\linewidth}
            \includegraphics[width=\linewidth]{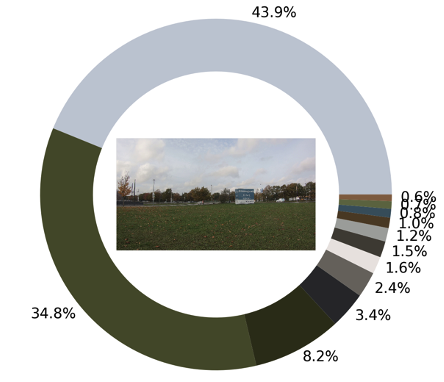}
            \caption{Scene 1 Ground}
          \end{minipage}
    \end{subfigure}
    \begin{subfigure}{\linewidth}
      \begin{minipage}{0.48\linewidth}
        \includegraphics[width=\linewidth]{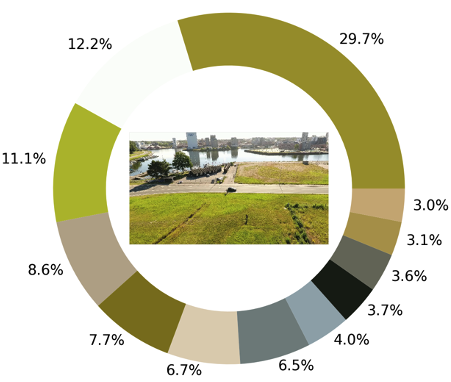}
        \caption{Scene 6 Aerial}
      \end{minipage}\hfill
      \begin{minipage}{0.48\linewidth}
        \includegraphics[width=\linewidth]{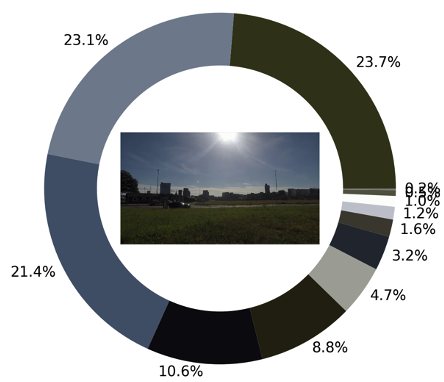}
        \caption{Scene 6 Ground}
      \end{minipage}
    \end{subfigure}
  \end{minipage}%
  \hfill
  \begin{minipage}[t]{0.48\textwidth}
    \begin{subfigure}{\linewidth}
      \begin{minipage}{0.48\linewidth}
        \includegraphics[width=\linewidth]{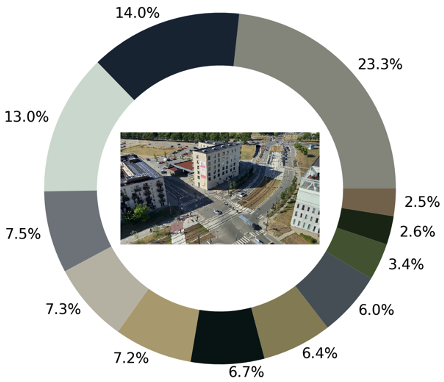}
        \caption{Scene 2 Aerial}
      \end{minipage}\hfill
      \begin{minipage}{0.49\linewidth}
        \includegraphics[width=\linewidth]{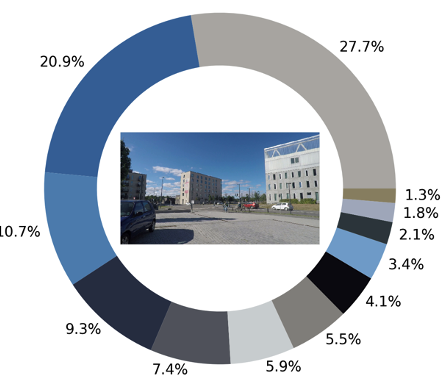}
        \caption{Scene 2 Ground}
      \end{minipage}
    \end{subfigure}
      \begin{subfigure}{\linewidth}
          \begin{minipage}{0.49\linewidth}
            \includegraphics[width=\linewidth]{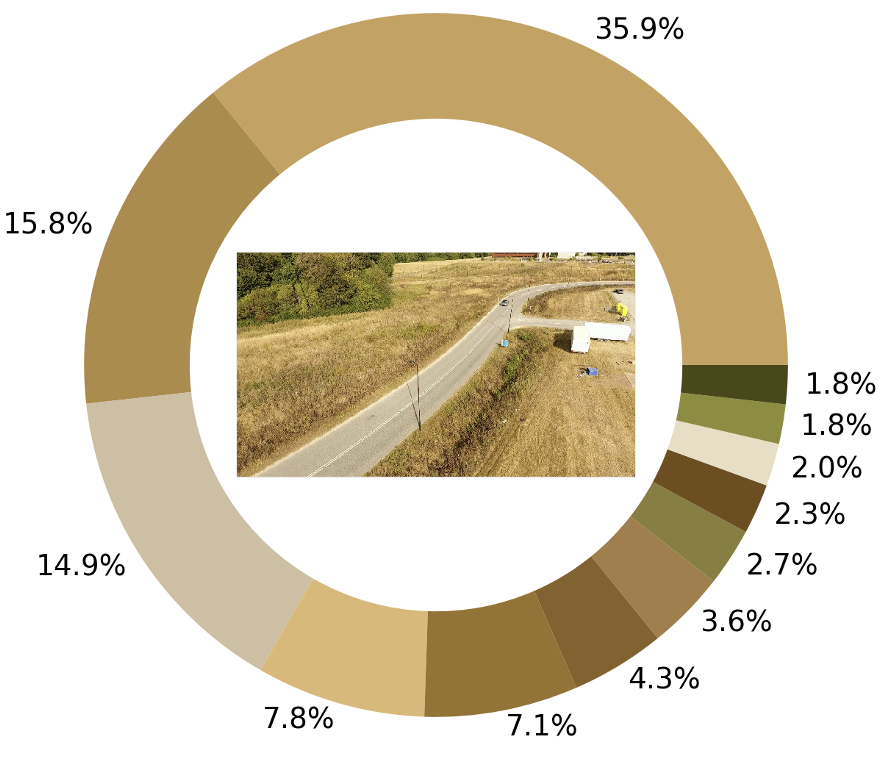}
            \caption{Scene 8 Aerial}
          \end{minipage}\hfill
          \begin{minipage}{0.49\linewidth}
            \includegraphics[width=\linewidth]{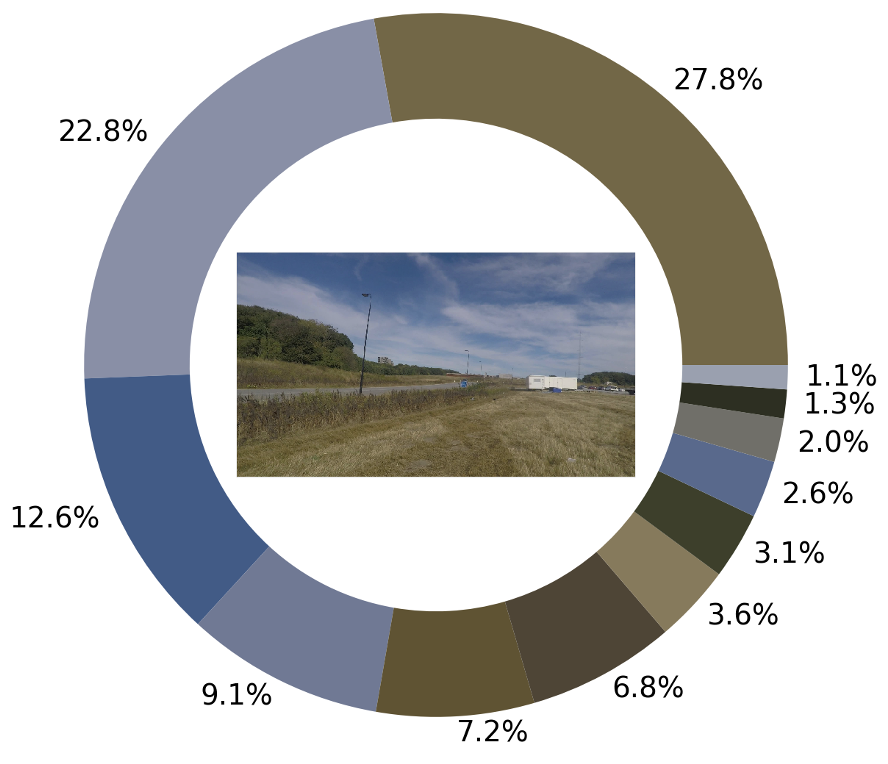}
            \caption{Scene 8 Ground}
          \end{minipage}
    \end{subfigure}
  \end{minipage}
  \caption{Dominant colors in different sample frames of MAVREC containing both views.}
  \label{fig:color_dominans_in_MAVREC}
\end{figure*}

\section{Addendum to the baseline and evaluation}\label{sec:appendix_baseline}

This section highlights the implementation details of our baseline DNN models; see Table \ref{table:DNN_models} and \ref{table:experiments_hyperparameters}. In Section \ref{sec:appendix_baseline_ablation}, we provide additional benchmarking results complementing Section \ref{sec:evaluation} in the main paper.

\subsection{Implementation details}\label{sec:appendix_implementation}
We train all object detectors for 39 epochs on ${600 \times 337}$ scaled images, except DETR.~DETR is a compute-heavy model and requires more than 39 training epochs \cite{DETR,D-DETR} for an optimal performance.~For supervised benchmarking, we train DETR with 100 object queries, and 10 classes (9 object class, 1 background class) for 300 epochs. For D-DETR, we used 900 queries and 20 classes. 
We adhere to the original training methodologies of the respective methods in order to train the object detectors specifically for the MAVREC dataset.


\smartparagraph{Computing environment.} For prototyping, we use a local testbed with an AMD EPYC 7501 32-Core Processor with 2.0GHz speed, 16 GB memory, and 1 Nvidia Tesla V100 GPU with 32 GB on-board memory. For training all the supervised baselines, we use two HPC nodes: \myNum{i} Node-1: 2x Intel(R) Xeon(R) Gold 6230 CPU with 2.10 GHz processing speed, 32 virtual cores, 192 GB memory, and 8 NVIDIA V100 GPU each with 32 GB on-board memory; \myNum{ii} Node-2: AMD EPYC 7F72 CPU with 3.2 GHz processing speed, 96 virtual cores, 2048 GB memory, and 8 NVIDIA A100 GPU each with 40 GB on-board memory. 
For training the semi-supervised baselines, we use a server with AMD EPYC 7662 CPU, 1024GB memory, 8 RTX A5000 GPU.

\begin{figure*}[ht]
  \begin{minipage}[t]{0.48\textwidth}
    \begin{subfigure}{\linewidth}
           \begin{minipage}{0.48\linewidth}
            \includegraphics[width=\linewidth]{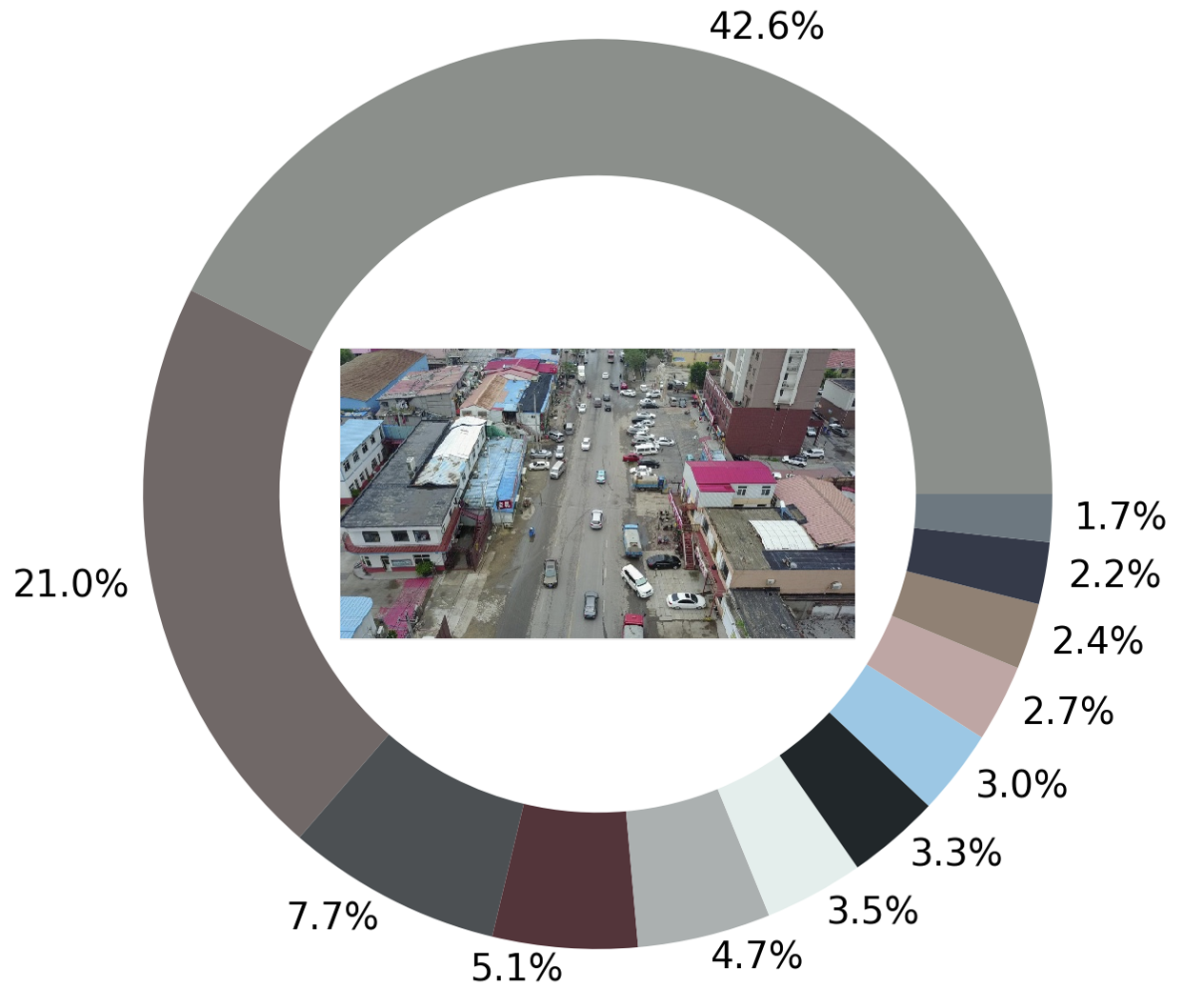}
            \caption{VisDrone DET}
          \end{minipage}\hfill
          \begin{minipage}{0.48\linewidth}
            \includegraphics[width=\linewidth]{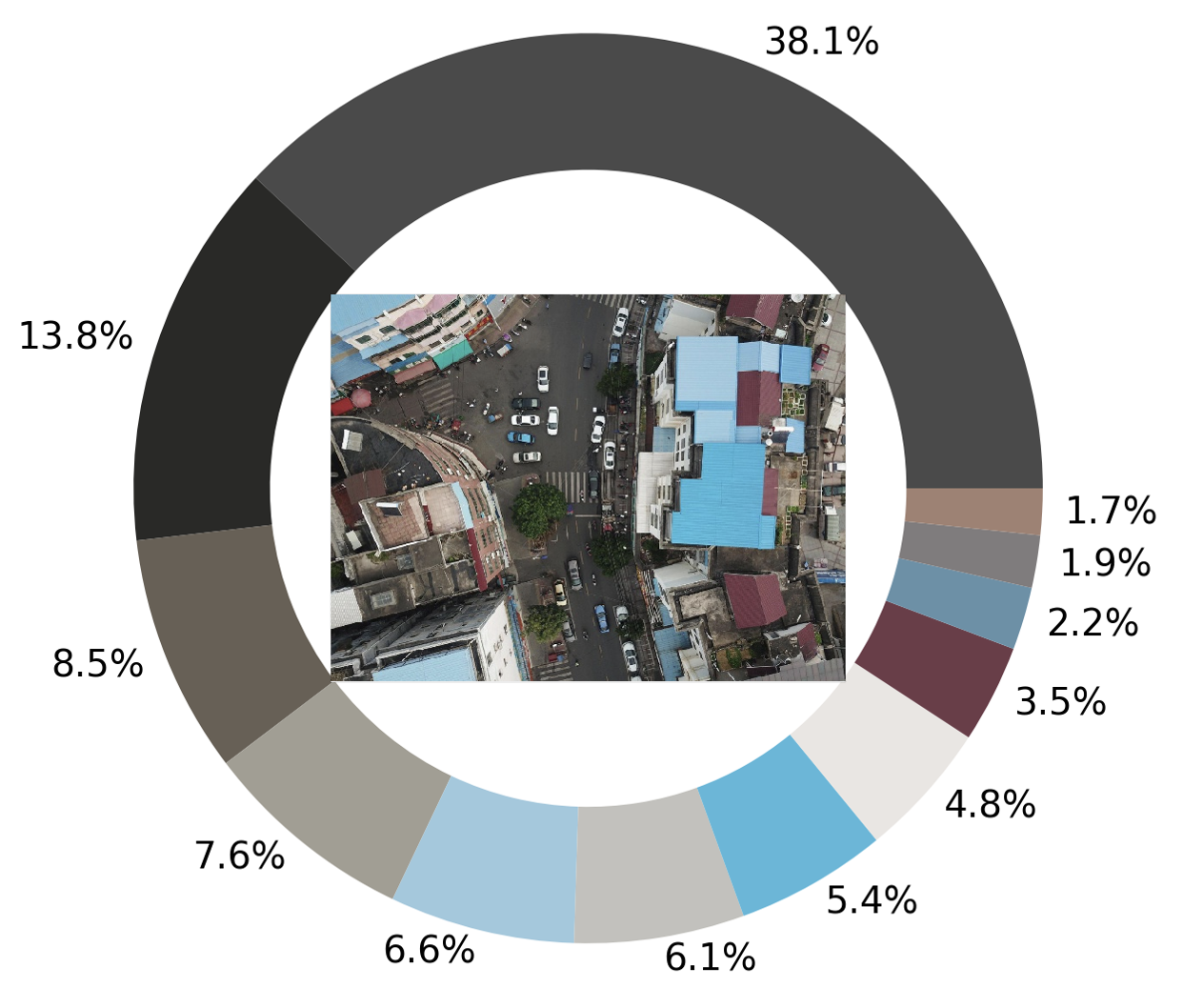}
            \caption{VisDrone DET}
          \end{minipage}
    \end{subfigure}
    \begin{subfigure}{\linewidth}
      \begin{minipage}{0.48\linewidth}
        \includegraphics[width=\linewidth]{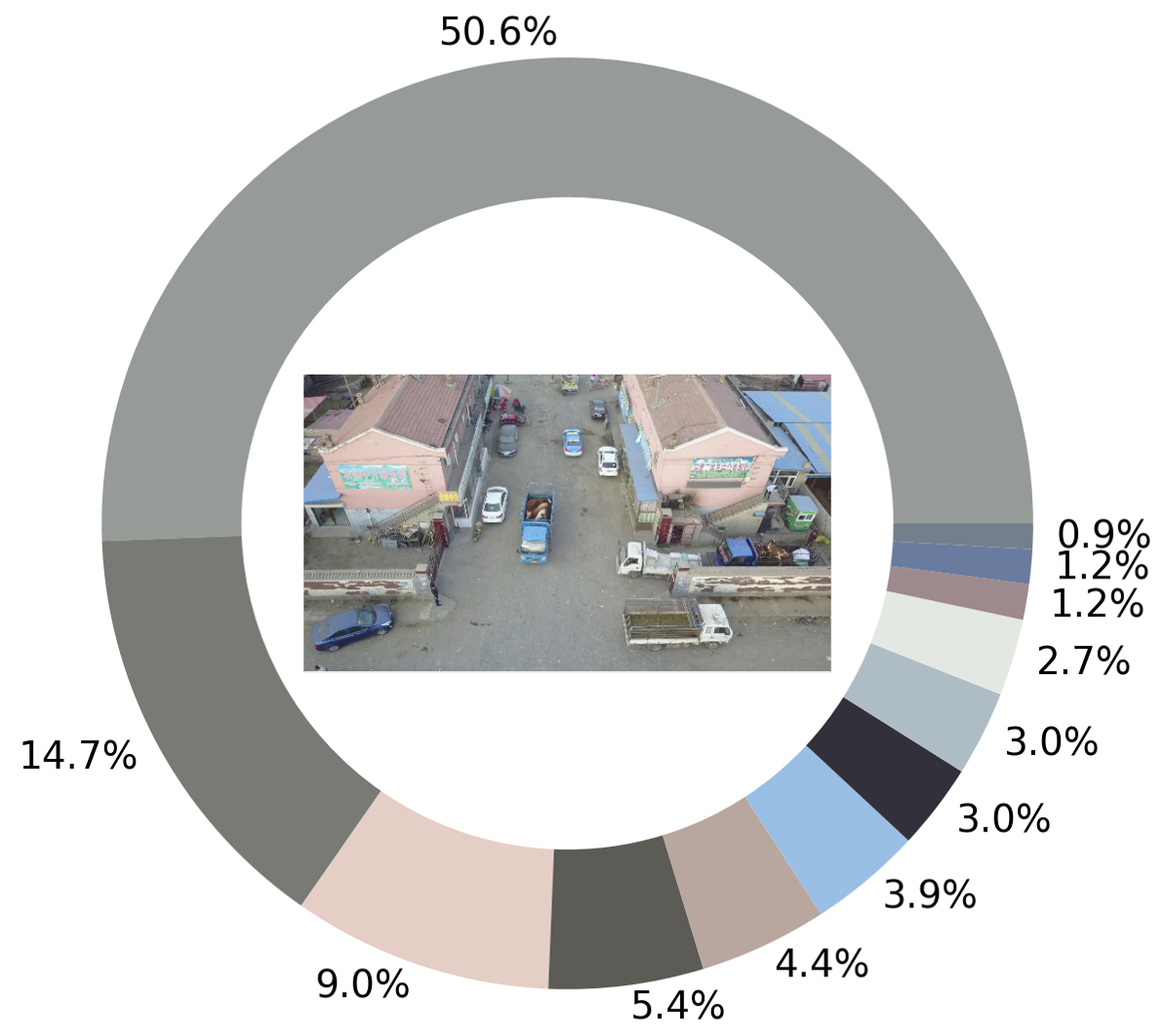}
        \caption{VisDrone MOT}
      \end{minipage}\hfill
      \begin{minipage}{0.48\linewidth}
        \includegraphics[width=\linewidth]{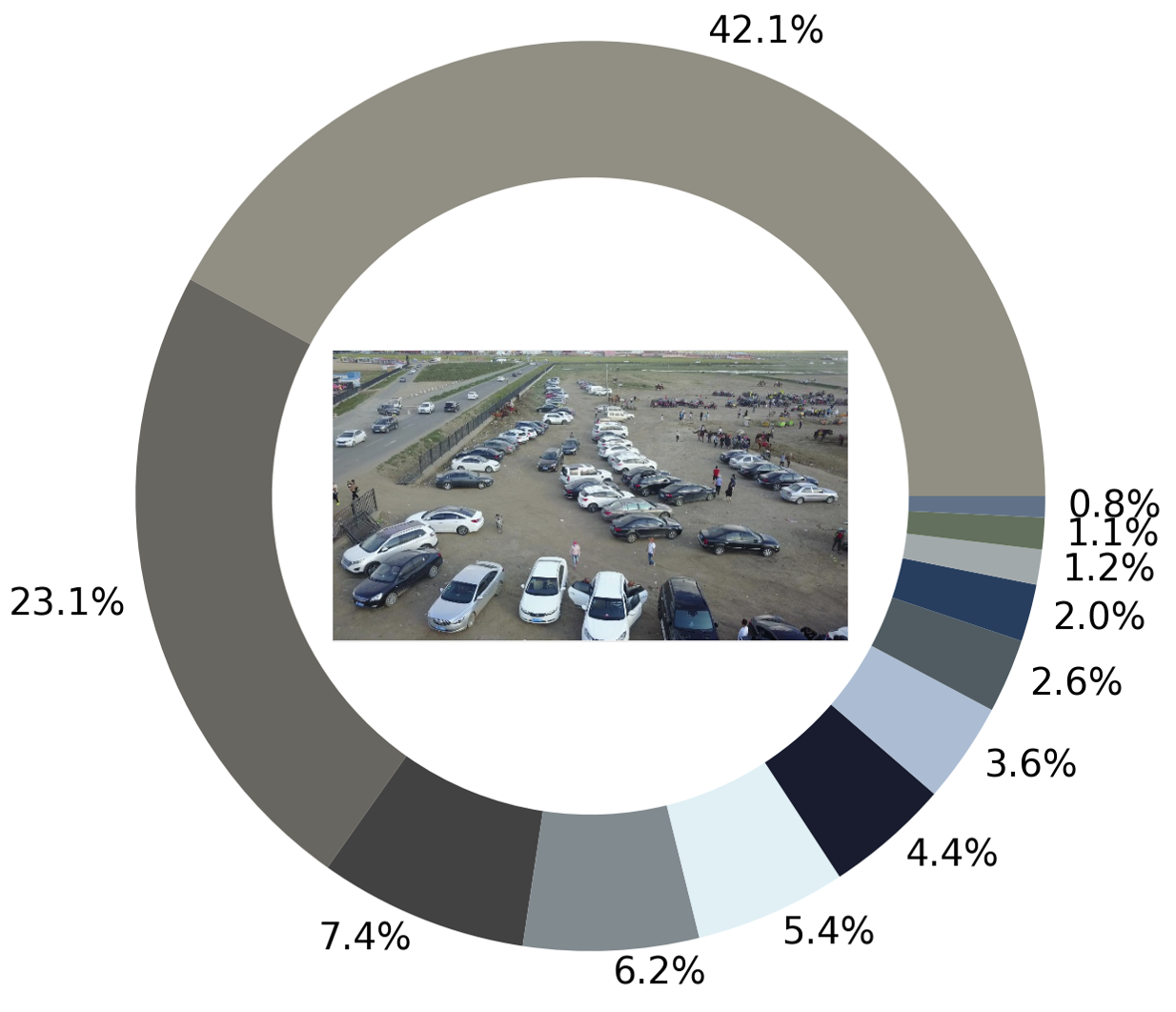}
        \caption{VisDrone MOT}
      \end{minipage}
    \end{subfigure}
  \end{minipage}%
  \hfill
  \begin{minipage}[t]{0.48\textwidth}
    \begin{subfigure}{\linewidth}
      \begin{minipage}{0.48\linewidth}
        \includegraphics[width=\linewidth]{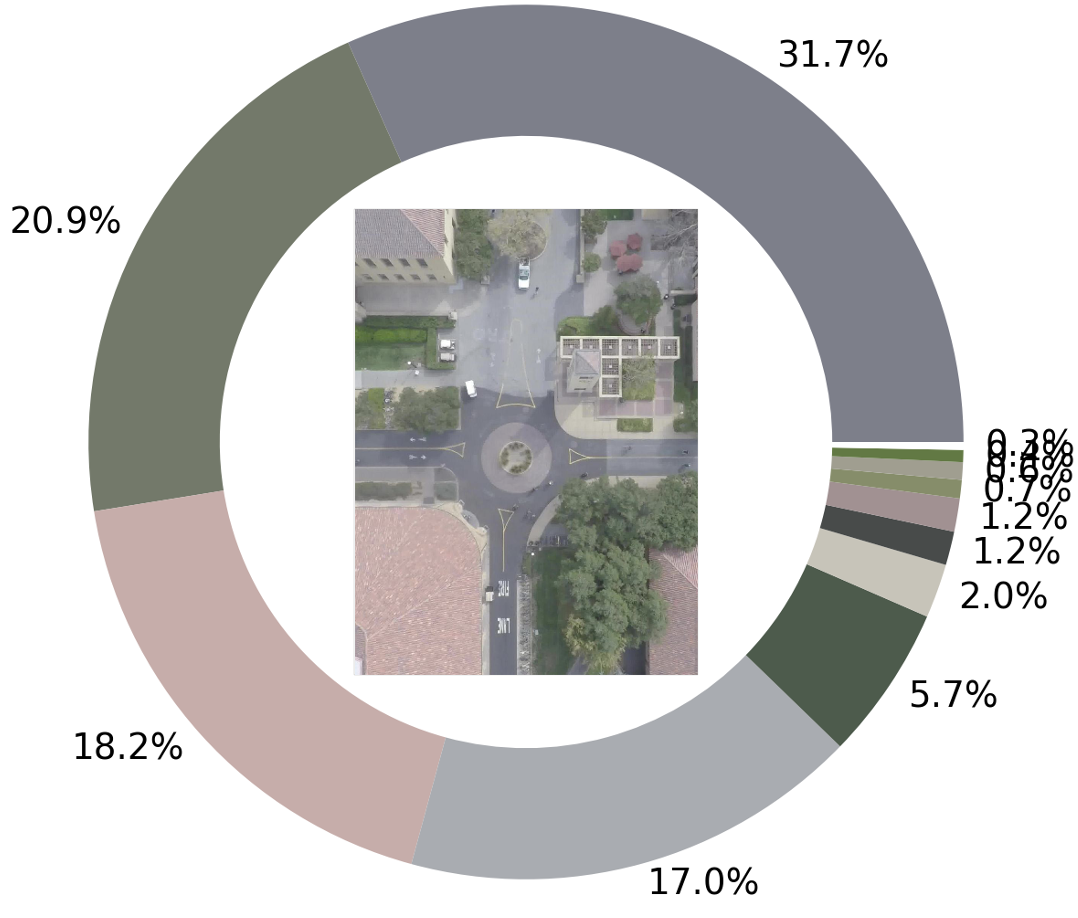}
        \caption{Campus}
      \end{minipage}\hfill
      \begin{minipage}{0.49\linewidth}
        \includegraphics[width=\linewidth]{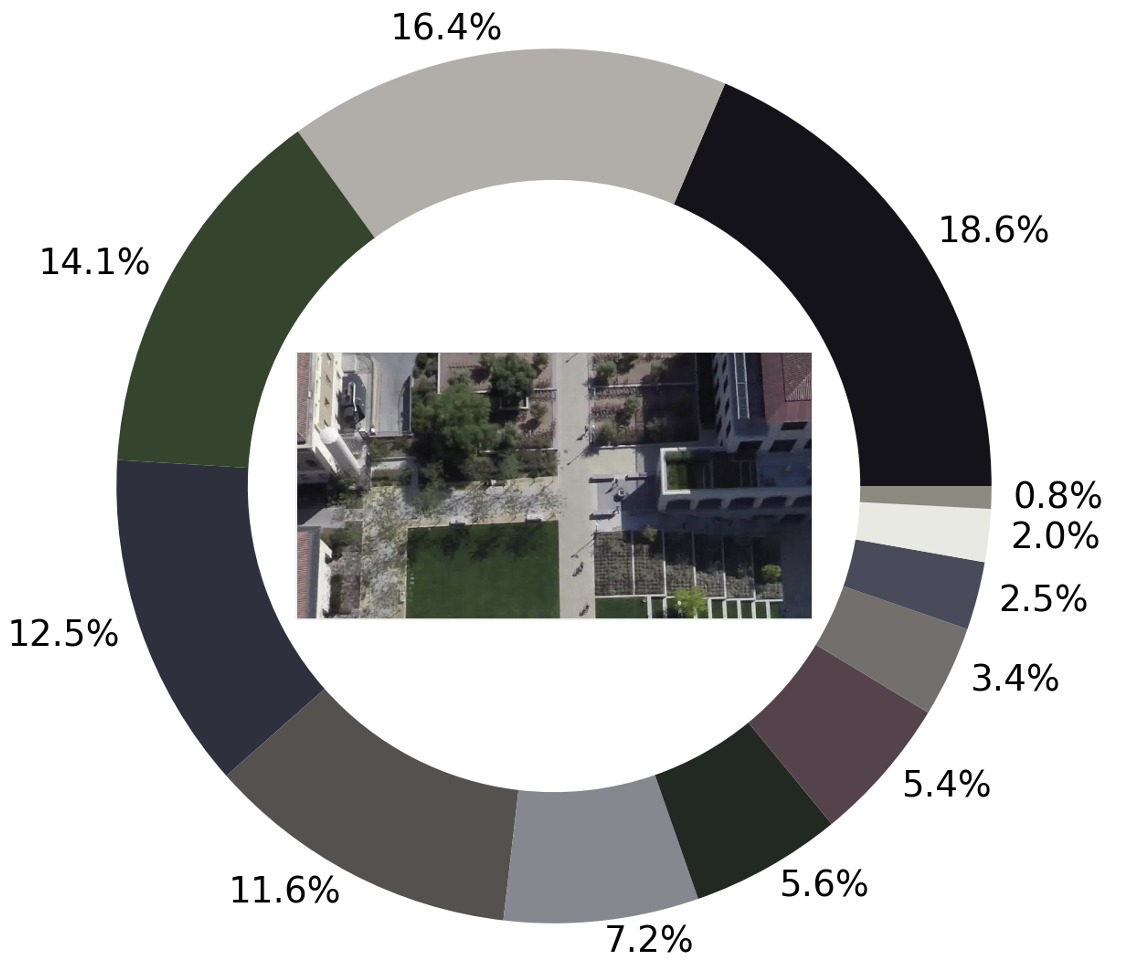}
        \caption{Campus}
      \end{minipage}
    \end{subfigure}
      \begin{subfigure}{\linewidth}
          \begin{minipage}{0.49\linewidth}
            \includegraphics[width=\linewidth]{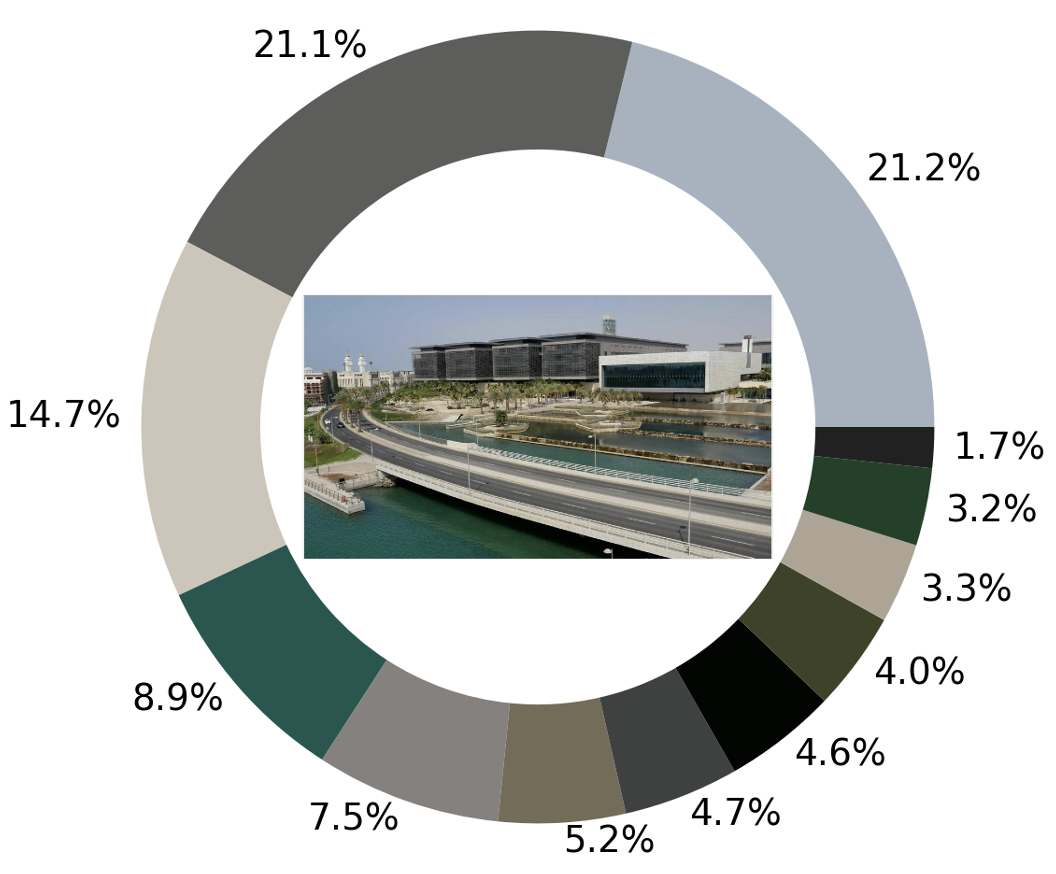}
            \caption{UAV123}
          \end{minipage}\hfill
          \begin{minipage}{0.49\linewidth}
            \includegraphics[width=\linewidth]{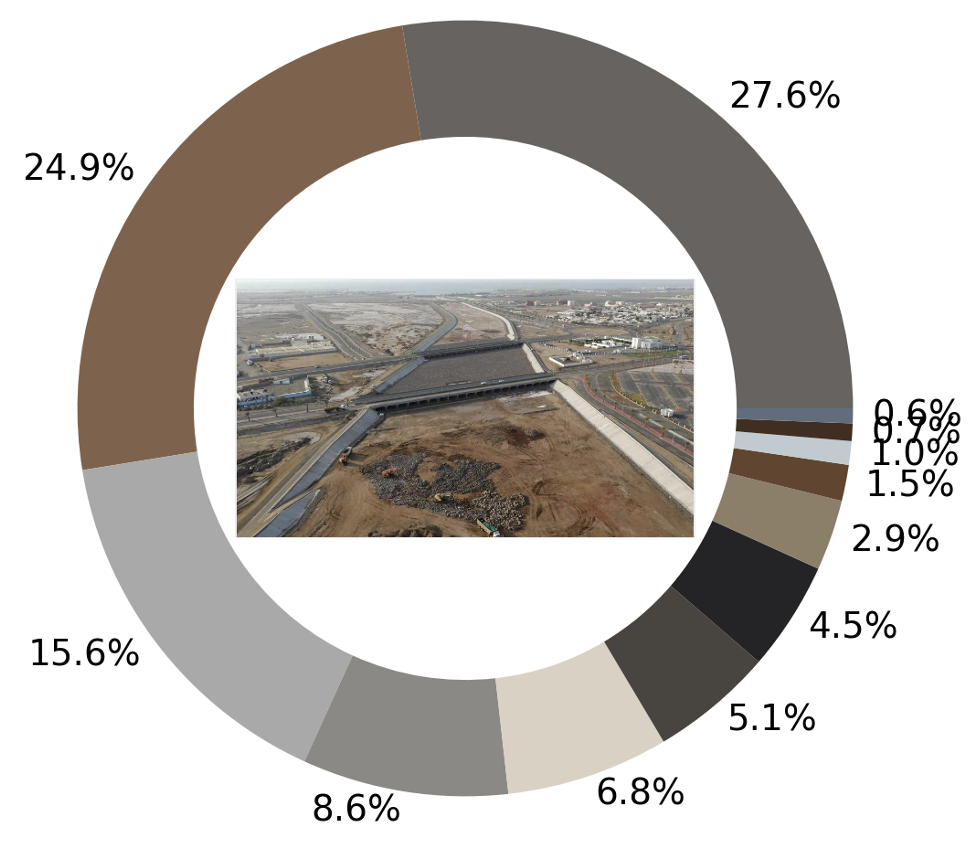}
            \caption{UAV123}
          \end{minipage}
    \end{subfigure}
  \end{minipage}
  \caption{Most dominant colors in the sample frames of VisDrone DET and MOT \cite{visdrone}, the Campus dataset \cite{campus}, and the UAV123 dataset \cite{UAV123}.}
  \label{fig:color_dominans_in_comparing_datsets}
\end{figure*}
\begin{table*}[t]
\small
\centering
\caption{\small{DNN models used for benchmarking. Note that $1M=10^6$.}}
\resizebox{\textwidth}{!}{ 
\begin{tabular}{cccccccc}
\midrule
\textbf{Type} & \textbf{Model} & \textbf{Task} & \textbf{Dataset} & \textbf{Parameters} & \textbf{Optimizer}  & \textbf{Platform} & \textbf{Metric} \\
\midrule
{CNN} & YoloV7  \cite{yolov7} & Detection  & MAVREC    & 36.5M & SGD-M \cite{nesterov2003introductory}  & PyTorch & mAP \\
\midrule
NAS & Yolo-NAS (L)  \cite{yolonas} & Detection  & MAVREC    & 51.1M & Adam \cite{kingma2014adam}  & PyTorch & mAP \\
\midrule
 & DETR~\cite{DETR} & Detection  & MAVREC   & 41M & Adam \cite{kingma2014adam}  & PyTorch & mAP  \\
Transformer & D-DETR~\cite{D-DETR} & Detection  & MAVREC and VisDrone  & 41M & Adam \cite{kingma2014adam}   & PyTorch & mAP  \\
&OMNI-DETR \cite{OMNI-DETR} & Detection  & MAVREC and VisDrone  & 41M & Adam \cite{kingma2014adam} & PyTorch & mAP  \\
 \midrule
\end{tabular}
}
\label{table:DNN_models}
\end{table*}

\begin{table*}[t]
\small
\centering
\caption{\small{Hyperparameters used for training each DNN model.}}
\resizebox{\textwidth}{!}{ 
\begin{tabular}{lcccccccc}
\midrule
    \textbf{Model} & \textbf{Backbone} & \textbf{Learning Rate} & \textbf{Batch Size} & \textbf{Weight Decay} & \textbf{Queries} & \textbf{Attention Heads}  & \textbf{Epochs} \\
    \midrule
    YoloV7  \cite{yolov7} & E-ELAN  & $1, 10^{-5}, 10^{-1}$ & 32    & $5 \times 10^{-4} $   & NA       & NA  &  39 \\
        Yolo-NAS (L) \cite{yolonas} & QA-RepVGG  & $10^{-6}$, $5 \times 10^{-4}$  & 16    & $10^{-4}$     & NA         & NA  &  39 \\
    \midrule
    DETR \cite{DETR} & ResNet50 \cite{he2016deep}  & $10^{-4}$  & 2 &  $10^{-4}$         & 100      & 16 & 300  \\
    D-DETR \cite{D-DETR} & ResNet50  & $2 \times 10^{-4}$  & 2 &  $10^{-4}$     & 900 & 16 & 39  \\
    OMNI-DETR \cite{OMNI-DETR} & ResNet50 & $10^{-4}$ & 2  & $10^{-4}$ & 900       & 16 &  39  \\
    \midrule
\end{tabular}
}
\label{table:experiments_hyperparameters}
\end{table*}

\subsection{Evaluation metric}\label{app:metric}
In this section we give brief description of the metric used in our experiments. 
\subsubsection{Average precision (AP)}
Average precision~(AP) is a standard metric for information retrieval tasks and is used for object detection and instance segmentation in computer vision. We pause here, and first explain the precision and recall of a model's performance in general. For a given test of predictions (of a model) and the corresponding ground-truth labels, the precision represents the proportion of correct class labels among all predicted positives. The recall represents the proportion of correct positive predictions among all actual positives. For an user-defined threshold, $t\in(0,1]$, denote precision as $P(t)$ and recall as $R(t)$ and are given as follows:
$$
    P(t) = \frac{\text{TP}}{\text{TP + FP}}\;\;\text{and}\;\;R(t) = \frac{\text{TP}}{\text{TP + FN}},
$$
where $TP, FP$, and $FN$ denote true positive, false positive, and false negative, respectively. The accuracy of the model's predictions is quantified by calculating the area under the precision-recall (PR) curve. 

In the context of object detection, next, we explain the intersection over union (IoU) metric. IoU describes the closeness of two bounding boxes (predicted and the ground truth) and is given as the ratio of the area  of intersection between the predicted box ($A_{\text{Predicted box}}$) and ground truth box ($A_{\text{Ground-truth box}}$) to that of their union: 
$$
    {\rm IoU} = \frac{A_{\text{Predicted box}} \cap A_{\text{Ground-truth box}}}{ A_{\text{Predicted box}} \cup A_{\text{Ground-truth box}}}.
$$
Naturally, IoU falls between 0 and 1, where 1 indicates a complete overlap between the two boxes and hence, perfect detection. While 0 indicates no overlap and hence, no detection. A detection box is assigned TP, FP, and FN based on the predicted label compared to the ground truth label and the IoU between the two boxes. In multi-class classification, the model outputs the conditional probability that the bounding box belongs to a certain object class. For a probability confidence threshold, $t\in(0,1]$, in general, the higher the number of detection, the lower the chances that the missed ground-truth labels, resulting in a higher recall. In contrast, the higher the confidence threshold, the more confident the model is its predictions, and this results in a higher precision. One can generate a PR curve based on different threshold values $t\in(0,1]$. Finally, the average precision (AP) is defined as the area under the PR curve:
$$
AP = \int_{t=0}^{1} p(t)dt. 
$$
In practice, numerical integration methods are used to approximately calculate this area. 

\smartparagraph{Mean average precision (mAP)} is the average AP across all object classes and is defined as follows:
\begin{center}
    ${\rm mAP}:= \frac{1}{|C|} \sum_{c \in C} AP_c,$
\end{center}
where $C$ is the set of all classes, $|C|$ is its the cardinality, and $AP_c$ be the AP for a class $c\in C$.

\subsubsection{COCO mAP \cite{COCO}}
Our results reported with the COCO mAP which is a cumulative sum of the average of multiple AP calculated at different IoU-thresholds ranging from $0.5$ to $0.95$ with an increment of $0.05$. COCO mAP is the average over 10 IoU levels on all classes.

\begin{figure*}
    \captionof{table}{{Supervised benchmark on aerial view of MAVREC (Validation Set). The first column indicates percentage of infused ground-view samples with the aerial-view train set. The last column indicates the relative change in mAP compared to the baseline model that is trained exclusively on aerial-view training set from MAVREC. {The top row represents training exclusively on aerial-view samples.}}}
  \begin{minipage}[t]{0.485\linewidth}
    \small
    \footnotesize 
    \centering
    \centering
    \begin{tabular}{cccccc}
    \midrule
    \textbf{Extra ground} & \texttt{AP} & ${\tt AP}_{50}$  & ${\tt AP}_{\tt S}$ & ${\tt AP}_{\tt M}$ &  \textbf{Relative}(\greenup\reddown)\\
         \textbf{view samples} &&&&& \textbf{change}~~~~~~~~\\
    \midrule
       { 0\%}  & {24.9} & {39.7} &	{27.6}  & {45.3} & {--}   \\
        12.5\%  & 34.4 & 63.8 &	31.6  & 64.3 & 162.6\% \greenup  \\
        25\%    & \textbf{48.5} & 73.3 &	45.8  & 73.6 &  270.2\% \greenup  \\
        37\%    & 44.4 & 71.0 &	41.9  & 71.9 &  238.9\% \greenup  \\
        50\%    & 40.8 & 69.0 &	38.6  & 73.5 &  211.5\% \greenup  \\
        75\%    & 44.2 & 66.6 &	40.8  & 79.5 &  237.4\% \greenup  \\
        100\%   & 42.3 & 65.7 &	38.9  & 68.4 &  222.9\% \greenup \\
    \midrule
    \end{tabular}\label{tab:baselines:supervised_street_ground_d_detr_validation}
    \subcaption{\footnotesize{D-DETR}}   
  \end{minipage}\hfill
  \begin{minipage}[t]{0.485\linewidth}
    \small
    \footnotesize 
    \centering
    \begin{tabular}{cccccc}
        \midrule
         \textbf{Extra ground} & \texttt{AP} & ${\tt AP}_{50}$  & ${\tt AP}_{\tt S}$ & ${\tt AP}_{\tt M}$ &  \textbf{Relative}(\greenup\reddown)\\
         \textbf{view samples} &&&&& \textbf{change}~~~~~~~~\\
        \midrule
         { 0\%}  & {31.3} & {57.7} &	{34.2}  & {61.2} & {--}   \\
          12.5\%  & 30.9 &	57.7 &	33.7  & 59.4 &  1.3\% \reddown \\
          25\%    & 31.4 &	58.1 &	34.3  & 65.9 &  0.3\% \greenup \\
          37\%    & 35.8 &	68.4 &	34.7  & 66.8 &  14.4\% \greenup \\
          50\%    & 30.9 &	58.2 &	33.7  & 62.2 &  1.3\%  \reddown\\
          75\%    & 45.3 &  79.1 &	43.0  & 79.6 &  44.9\% \greenup \\
          100\%   & \textbf{48.3} &	78.6 &	43.0  & 85.0 &  54.5\% \greenup\\
        \midrule
    \end{tabular}\label{tab:baselines:supervised_street_ground_yolo_validation}
    \subcaption{\footnotesize{YoloV7}}
  \end{minipage}
  \end{figure*}\label{tab:baselines:supervised_street_ground_validation}
\begin{figure*}
    \captionof{table}{\small{Supervised benchmark on aerial view of MAVREC (Test Set). The first column indicates percentage of infused ground-view samples with the aerial-view train set. The last column indicates the relative change in mAP compared to the baseline model that is trained exclusively on aerial-view training set from MAVREC.}}
  \begin{minipage}[t]{0.485\linewidth}
    \small
   \footnotesize 
    \centering
    \centering
    \begin{tabular}{cccccc}
    \midrule
    \textbf{Extra ground} & \texttt{AP} & ${\tt AP}_{50}$  & ${\tt AP}_{\tt S}$ & ${\tt AP}_{\tt M}$ &  \textbf{Relative}(\greenup\reddown)\\
         \textbf{view samples} &&&&& \textbf{change}~~~~~~~~\\
    \midrule
        12.5\%  & 39.8 & 68.6 &	39.9  & 55.8 & 286.4\% \greenup  \\
        25\%    & \textbf{44.8} & \textbf{71.5} &	\textbf{42.9}  & \textbf{72.4} & 335.0\% \greenup  \\
        37\%    & 41.1 & 69.1 &	39.7  & 61.6 & 299.0\% \greenup  \\
        50\%    & 36.0 & 65.8 &	33.0  & 54.1 & 249.5\% \greenup  \\
        75\%    & 28.7 & 56.6 &	26.6  & 62.8 & 178.6\% \greenup  \\
        100\%   & 39.9 & 65.8 &	32.5 & 70.6 & 287.4\% \greenup \\
    \midrule
    \end{tabular}\label{tab:baselines:supervised_street_ground_d_detr_test}
    \subcaption{\footnotesize{D-DETR}}   
  \end{minipage}\hfill
  \begin{minipage}[t]{0.485\linewidth}
    \small
   \footnotesize
    \centering
    \begin{tabular}{cccccc}
        \midrule
         \textbf{Extra ground} & \texttt{AP} & ${\tt AP}_{50}$  & ${\tt AP}_{\tt S}$ & ${\tt AP}_{\tt M}$ &  \textbf{Relative}(\greenup\reddown)\\
         \textbf{view samples} &&&&& \textbf{change}~~~~~~~~\\
        \midrule
          12.5\%  & 29.5 & 55.6 &	28.8  & 64.6 & 5.6\% \reddown \\
          25\%    & 30.1 & 56.2 &	29.5  & 64.1 & 3.8\% \reddown \\
          37\%    & 33.1 & 63.3 &	30.4  & 70.0 & 5.8\% \greenup \\
          50\%    & 29.6 & 59.0 &	29.2  & 66.1 & 5.4\%  \reddown\\
          75\%    & 40.5 & 74.6 &	36.7  & 74.7 & 29.4\% \greenup \\
          100\%   & \textbf{45.5} & \textbf{76.1} &	\textbf{43.8}  & \textbf{81.6} & 45.4\% \greenup\\
        \midrule
    \end{tabular}\label{tab:baselines:supervised_street_ground_yolo_test}
    \subcaption{\footnotesize{YoloV7}}
  \end{minipage}
  \end{figure*}\label{tab:10}
\begin{table*}
\small
\centering
\caption{\small{Mix-up benchmarks after 39 epochs; the test perspective is the aerial view.}}
\small
\centering
\begin{tabular}{ccc cccc ccc}
\midrule
\multirow{2}{*}{\textbf{Model}} & \multirow{2}{*}{\textbf{Mix-up parameter}} & \multicolumn{4}{c}{\textbf{Validation Set}} & \multicolumn{4}{c}{\textbf{Test Set}} \\
\cmidrule(rl){3-6} \cmidrule(rl){7-10}
& & {\tt AP} & ${\tt AP}_{50}$  & ${\tt AP}_{S}$ & ${\tt AP}_{M}$  & {\tt AP} & ${\tt AP}_{50}$  & ${\tt AP}_{S}$ & ${\tt AP}_{M}$\\

\midrule                      
                           &$[0.65, 1.0]$  & 22.8         & 44.0 & 22.6 & 49.8 & 22.3 & 42.4 & 22.0 & 50.1 \\
                          &$[0.75, 1.0]$   & \textbf{33.4} & 56.0 & \textbf{31.2} & 56.1 & \textbf{29.1} & 49.6 & 27.0 & 47.7\\
       D-DETR            & $[0.85, 1.0]$   & 28.2 & 50.1 & 25.5 & 55.9 & 23.5 & 44.9 & 22.0 & 44.9\\
                         & $0.9$           & 25.8 & 41.6 & 28.3 & 46.4 & 23.3 & 41.3 &  25.0 & 42.3 \\
                        & $[0.0, 1.0]$     & 6.4 & 12.5 & 8.7 & 9.1 & 10.4 & 17.7 & 12.9 & 13.3\\
\midrule

YoloV7 & $[0.75, 1.0]$  & {30.3} & \textbf{58.6} & {29.8} & \textbf{60.7} & 28.5 & \textbf{55.3} & \textbf{27.9} & \textbf{57.9}\\
\midrule
\end{tabular}\label{table:ablation_mixup_DDETR}
\end{table*}

\begin{figure*}[t]
     \centering
     \includegraphics[width=\textwidth]{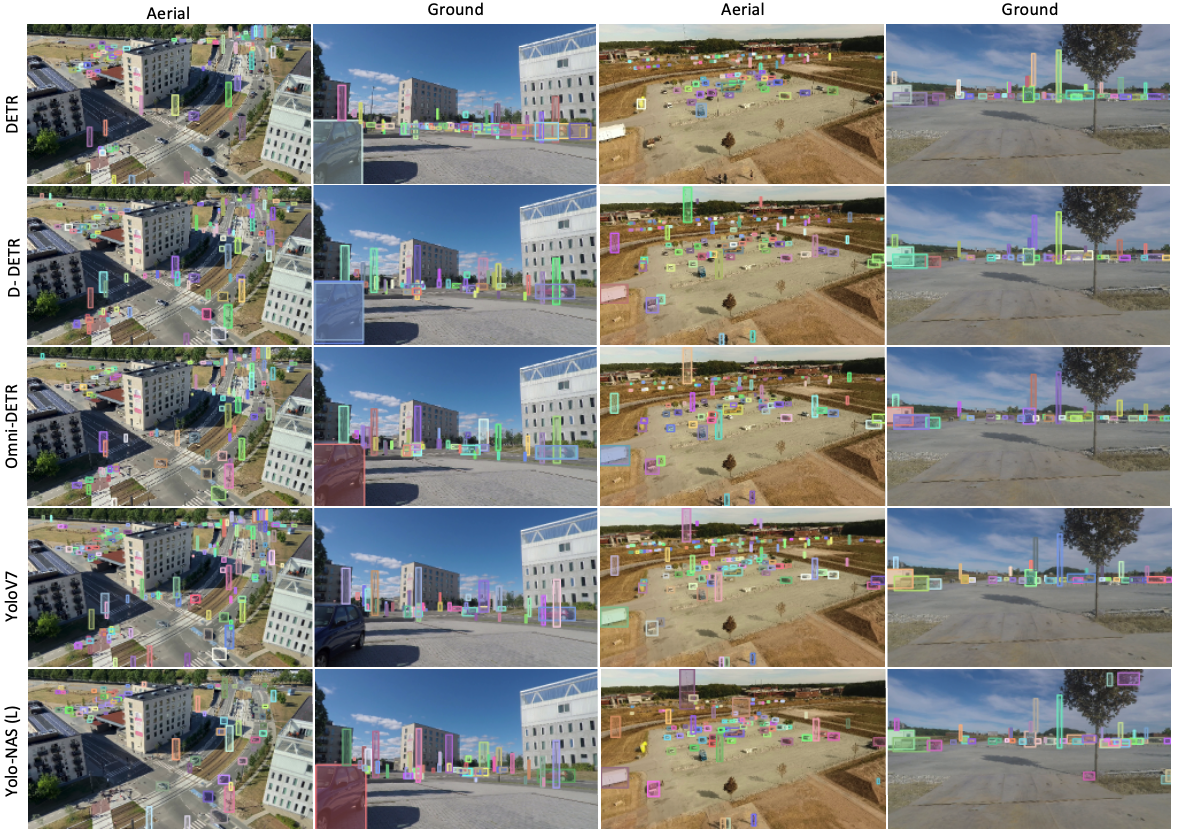}
     \caption{Qualitative inference results of different DNN models on the test set of MAVREC.}
     \label{Figure:qualitative_results}
\end{figure*}

\subsection{Additional baseline results}\label{sec:appendix_baseline_ablation}
In Table \ref{tab:baselines:supervised_street_ground_d_detr_test}, we provide the supervised benchmark results on the test of the aerial-view of MAVREC by using D-DETR and YoloV7. Except a few minor discrepancies, overall our observation in the main paper holds on MAVREC test set results --- We demonstrate that the inclusion of ground-view samples substantially improves the object detection performance. 

\subsubsection{Benchmarking with mix-up across views}\label{sec:appendix_mixup}
We use the mix-up strategy to naturally augment and combine the dual views of our data.

\smartparagraph{Why mix-up?} Previously, we demonstrated that jointly training the aerial-view samples with ground-view samples substantially improves object detection from an aerial perspective; see Section \ref{sec:supervised-benchmark}. Nevertheless, a natural question could be---Can a {\em data-augmentation strategy} be able to improve the aerial-visual perception while aerial-view images are {\em augmented} with corresponding ground-view images? This motivates us to use mix-up \cite{zhang2018mixup} as an augmentation strategy that can combine these two views. 

The mix-up is a data augmentation technique that creates a convex combination of the input data pair and their labels and reduces the inductive bias \cite{zhang2018mixup}. For input pair, ${(x_{A}, x_{G})},$ and their corresponding labels, $(y_{A}, y_{G})$, mix-up creates new input, ${x_{m} = \lambda x_{A} + (1-\lambda) x_{G},}$ and label, ${y_{m} = \lambda y_{A} + (1-\lambda) y_{G},}$ where ${\lambda\in[0,1]}$ is the mixing parameter sampled from a $\beta_{\alpha,\beta}$-distribution with $\alpha=\beta=1$. Thus, we apply mix-up to the 8605 pairs of aerial and ground-view samples in the input space, while the testing perspective remains the aerial view. Note that our approach to mix-up differs from the original concept. We consistently apply mix-up across the views for the same samples, as opposed to performing mix-up among random samples within a batch.

\smartparagraph{D-DETR and YoloV7 training results with mix-up.} Each sample, $S$, consists of a pair of ground and aerial images, $(x_G,x_A)$ of the same scene. During training, we sample the mixing parameter, $\lambda\sim \beta_{1,1}$ such that ${\lambda>0.5}$, resulting in $A$ as the dominant image. The best mAP corresponds to $\lambda\in[0.75,1]$ for D-DETR on MAVREC; see Table \ref{table:ablation_mixup_DDETR} for ablation study for the optimal $\lambda$. For YoloV7, we use the best $\lambda$ from the mix-up D-DETR experiments. The results in Table \ref{table:ablation_mixup_DDETR} suggest that D-DETR with mix-up parameter $\lambda>0.5$ renders a better performance than vanilla D-DETR trained only on aerial view images; see Table \ref{tab:baselines:supervised_benchmark} in Section \ref{sec:evaluation}. YoloV7 with mix-up parameter, $\lambda\in[0.75,1]$ performs better than the mix-up D-DETR. Overall, we can conclude that mix-up D-DETR is better than the vanilla D-DETR model trained only on aerial images; for YoloV7, the performance is almost similar. In our experiments, mix-up technique uses 17,210 images (8,605 pairs of ground and aerial view images), while only {\em a fraction of the 8,605 ground view images} jointly trained with 8,605 aerial images can surpass its performance as evident from Tables \ref{tab:baselines:supervised_street_ground_d_detr_validation} and \ref{tab:baselines:supervised_street_ground_d_detr_test}. In conclusion, although our cross-view mix-up technique enhances object detection performance, the superior strategy for improving aerial detection performance is to train aerial-view samples together with ground-view samples. Future work will explore combining both the strategies (joint training and mix-up) to improve the performance of downstream tasks in aerial perspective.


\section{Reproducibility, privacy, {safety,} and broader impact}\label{sec:appendix_impact}
This paper introduces a large-scale, high-definition ground and aerial-view video dataset, \textbf{MAVREC}, and performs extensive benchmarking on the data. The dataset is open-source, fully curated, prepared,
and we plan to release our dataset via an academic website for research, academic, and commercial use. The dataset is protected under the CC-BY license of creative commons, which allows the users to distribute, remix, adapt, and build upon the material in any medium or format, as long as the creator is attributed. The license allows MAVREC for commercial use. As the authors of this manuscript and collectors of this dataset, we reserve the right to distribute the data. Additionally, we provide the code, data, and instructions needed to reproduce the main experimental baseline results, and the statistics pertinent to the dataset. We specify all the training details (e.g., data splits, hyperparameters, model-specific implementation details, compute resources used, etc.). 

{We conduct the recording in public spaces in compliance with the European Union's drone regulations. In Scandinavian countries, video recording falls under surveillance if the recording lasts continuously over 6 hours; our recorded clips are only a few minutes long. Moreover, in crowded intersections, to adhere to the drone-safety protocols, we did not operate drones, instead, we used user-grade handheld cameras from a high riser. As our recordings follow these protocols, the university's legal team confirmed that we do not need additional permissions for our data collection process or publication.}

{MAVREC is a traffic-centric dataset, with repetitive human activities limited to bicycling, stopping at red traffic lights, and occasionally walking by. The position and distance of the ground and drone cameras do not allow any explicit human recognition.} There are {many} human subjects present in the data, although there are no personal data that can resemble shreds of evidence, reveal identification, or show offensive content. 
{By watching the video clips from the MAVREC, the university's legal experts have concluded that the MAVREC does not have recognizable human subjects and hence does not interfere with privacy.} Therefore, MAVREC is not subject to IRB (for North America) or GDPR (for Europe) compliance as it has no privacy concerns. We thoroughly discussed and validated this issue with appropriate legal experts. 

The dataset can be used by multiple domain experts. Its application includes but is not only limited to surveillance, autonomous driving \cite{Marathe_2022_CVPR, bojarski2016end}, robotics and instructional videos \cite{wu2022_survey}, environmental monitoring \cite{PanETAL2021}, heavy industrial infrastructure inspection \cite{ASK2020}, developing livable and safe communities \cite{Islam2023EffectOS, cityscape, UCF-SST-CitySim}, and a few to mention. Although we do not find any foreseeable harms that the dataset can pose to human society, it is always possible that some individual or an organization can use this idea to devise a {\em technique} that can appear harmful to society and can have evil consequences. However, as authors, we are absolutely against any detrimental usage of this dataset, regardless by an individual or an organization, under profit or non-profitable motivation, and pledge not to support any detrimental endeavors concerning our data or the idea therein.

{
\subsection{Maintenance plan}
The authors are responsible for maintenance and continuous hosting of the dataset on the web. The project lead will assign a research assistant for this purpose.  For any queries regarding corrections, annotations and learning algorithm the user can reach the maintenance team at \texttt{MAVRECdataset@gmail.com.}

The authors will release the subsequent versions of the dataset to address any reported errors and incorporate proper corrections. The authors will also add annotations if any and delete faulty annotations. The authors will determine the necessity for these updates annually, and subsequently, the latest version will be published on the website along with all previous versions. Retaining access to earlier versions of the  dataset would allow the users for reference during their evaluations and verify their results with the proper versions. To differentiate between the versions, each version will be assigned a unique number.}

{
\section{Motivation for research challenges on MAVREC dataset} 
We offer the research community object detection challenges to investigate through a synchronized multi-view dataset. We also encourage the researchers to exploit how a multi-view dataset (with partial annotation) can provide the basis for developing techniques to improve performance in aerial object detection. We highlight a few challenges below: 
\begin{enumerate}
    \item Utilizing the synchronized views and the temporal dimension not provides implicit information and offers a resource-efficient way to enhance performance using unsupervised and semi-supervised techniques. Resource-heavy recording setup or annotations is not required to accomplish this. An advancement in this direction would bring a new era of research in an area increasingly driven by large amounts of data.
    \item We underline the need for future research in sampling optimally aerial and ground views. This extends not only to MAVREC but also to other datasets from different domains and modalities. The insights gained from such research could serve as a cornerstone for comprehending more optimal dataset constituents that contribute to DNN's perception. Further, the research community can discover ways to identify samples that foster this understanding and those that hinder it. 
    \item Recovering objects from one view using the other has multiple motivations: \myNum{i} training a model on one of the views encourages us to develop techniques that can act as a backup to sensor failure in another view. This can have multiple practical use cases in surveillance and robotics. \myNum{ii} Recovering objects from an easier learned view can aid learning of a much more difficult view by information transfer between these two views. Encouraging such algorithms would further promote mapping between the views without sophisticated systems such as global navigation satellite/inertial navigation systems~(GNSS/INS). 
\end{enumerate}

}


\end{document}